\documentclass[10pt,twocolumn,letterpaper]{article}

\usepackage{cvpr}              
\definecolor{cvprblue}{rgb}{0.21,0.49,0.74}
\usepackage[pagebackref,breaklinks,colorlinks,allcolors=cvprblue]{hyperref}
\usepackage{hyperref}
\usepackage{url}
\usepackage[table]{xcolor}  
\usepackage{tabularx}
\usepackage{float}
\usepackage{pifont}
\usepackage{soul} 
\usepackage{color, xcolor} 
\usepackage[dvipsnames,svgnames]{xcolor}
\usepackage{tabularx}
\usepackage{graphicx}
\usepackage{amsmath}
\usepackage{array}
\usepackage{amssymb}
\usepackage{booktabs}
\usepackage{multirow}
\usepackage{makecell}
\usepackage{caption}
\usepackage{xspace}
\usepackage{wrapfig}
\usepackage{booktabs} 
\usepackage{caption}  
\usepackage{subcaption}
\usepackage{booktabs}
\usepackage{pifont}
\usepackage{xcolor}
\usepackage{colortbl}

\usepackage{titletoc}
\usepackage{xurl}
\def\paperID{} 
\def\confName{CVPR}
\def\confYear{2026}

\newcommand{\ours}{\textsc{HumanVBench}\xspace}

\title{\ours: Probing Human-Centric Video Understanding in MLLMs with Automatically Synthesized Benchmarks}

\author{Ting Zhou$^{1}$, Daoyuan Chen$^{2}$, Qirui Jiao$^{1}$, Bolin Ding$^{2}$, Yaliang Li$^{2}$, Ying Shen$^{1,3,4 \ast}$\\
~\\
$^{1}$Sun Yat-Sen University, $^{2}$Alibaba Group, $^{3}$ Peng Cheng Laboratory, \\
$^{4}$ Guangdong Provincial Key Laboratory of Fire Science and Intelligent Emergency Technology
\\
\small\texttt{\{zhout88,jiaoqr3\}@mail2.sysu.edu.cn, sheny76@mail.sysu.edu.cn} \\
\small\texttt{\{daoyuanchen.cdy,bolin.ding,yaliang.li\}@alibaba-inc.com}\\
}

\begin{document}

\twocolumn[{
\maketitle
\begin{center}
    \includegraphics[width=\textwidth]{HCBench_fig/fig1.pdf}
    \captionof{figure}{\ours is a benchmark for human-centric video understanding that features 16 fine-grained tasks, each denoted by its acronym and the number of included QA instances (middle blue box). It is synthesized with minimal human intervention by using a Video Annotation Pipeline (upper left, purple box) and a Distractor-Included QA Synthesis Pipeline (lower left, green box).}

    \label{fig:HumanVBench_fig}
\end{center}
}]

\insert\footins{\footnotesize $^\ast$Corresponding author.}

\begin{abstract}

Evaluating the nuanced human-centric video understanding capabilities of Multimodal Large Language Models (MLLMs) remains a great challenge, as existing benchmarks often overlook the intricacies of emotion, behavior, and cross-modal alignment. We introduce \ours, a comprehensive video benchmark designed to rigorously probe these capabilities across 16 fine-grained tasks. A cornerstone of our work is a novel and scalable benchmark construction methodology, featuring two automated pipelines that synthesize high-quality video annotations and challenging multiple-choice questions with minimal human labor. By leveraging state-of-the-art models for annotation and systematically converting model-induced errors into plausible distractors, our framework provides a generalizable ``machine'' for creating nuanced evaluation suites. Our extensive evaluation of 30 leading MLLMs on \ours reveals critical deficiencies, particularly in perceiving subtle emotions and aligning speech with visual cues, with even top proprietary models falling short of human performance. We open-source \ours and our synthesis pipelines to catalyze the development of more socially intelligent and capable video MLLMs.
\end{abstract}  
\vspace{-1em}
\section{Introduction}
In recent years, Multimodal Large Language Models have emerged as pivotal advancements, extending traditional language models to process diverse modalities such as text, images, and videos \citep{qin2024synergy,DBLP:journals/csur/KuangSXLXLLCLH25, jiao2025imgdiff,jiao2024adaptivetraining, DBLP:conf/acl/KuangL0L0J25, jiao2025detailmaster}. Video-oriented MLLMs, in particular, have drawn growing interest for their potential to interpret video content in ways closely aligned with human perception \citep{tang2025video}. However, the extent to which these models truly achieve human-like understanding, especially in complex human-centric scenarios, remains an open question.

Human-centric scenes in videos are rich in emotional, behavioral, and interactive content, requiring robust foundational human perceptual skills from video understanding models. Existing MLLM benchmarks mainly target general content comprehension, but often overlooking the fine-grained and multi-dimensional evaluation of essential human perceptual abilities, like nuanced understanding of emotions, behaviors, and person recognition. Synchronizing speech with visual cues, a key aspect of human perception, is also frequently neglected. Unlike humans who easily detect audio-visual mismatches, models struggle with speaker identification and lip-sync alignment. 
These foundational skills underpin advanced human-related reasoning tasks, including narrative reasoning, intention inference, and social intelligence. Therefore, a foundational human-centric video understanding benchmark is necessary, which requires not only new evaluation dataset but also scalable and systematic frameworks for its creation.

To address this, we introduce a novel and scalable framework for automated video benchmark creation. This framework is powered by two core pipelines: 1) the Human-Centric Video Annotation Pipeline, which leverages over twenty state-of-the-art (SOTA) data processing operators to produce dense, multimodal, automated annotations;
and 2) the Distractor-Included QA Synthesis Pipeline, which generates high-quality multiple-choice questions with semantically plausible distractors. As a powerful instantiation of our framework, we present HumanVBench, a pioneering benchmark featuring 16 fine-grained QA tasks specifically tailored for human-centric analysis in video MLLMs. These tasks are organized around two core dimensions: inner emotion and outer manifestation, as depicted in Figure  \ref{fig:HumanVBench_fig}.

A key innovation of our framework lies in its model-driven design. Unlike conventional benchmarks that rely on laborious human annotation, our approach automates the multi-modal, fine-grained annotation process by integrating diverse task-specific algorithms and models. The Distractor-Included QA Synthesis Pipeline further ensures the challenging nature of the questions by leveraging multi-model ensembles and their common failure modes to generate semantically deceptive distractors. The strategy transforms the human role from manual creation to efficient quality review, significantly boosting both construction efficiency and scalability. Moreover, our method is easily adaptable to  ``in-the-wild'' video data, enabling benchmarks creation beyond controlled or domain-specific settings.

Through \ours, we comprehensively evaluate 30 state-of-the-art video MLLMs, covering both open-source and commercial models. Our results reveal notable gaps between current model capabilities and human-level video understanding, particularly in nuanced emotion perception. While the proprietary models demonstrate closer human-like accuracy, open-source models often misclassify emotions due to temporal noise, underscoring the need for further architectural improvements and refined datasets.

In summary, our contributions are as follows:

\textbullet\ We introduce \ours, a novel video benchmark for MLLMs that emphasizes fine-grained human comprehension in videos, focusing on emotion perception, person recognition, behavioral analysis, and speech-visual alignment.

\textbullet\ We propose a novel and scalable framework for automated video benchmark construction, featuring two advanced pipelines for multi-modal annotation and distractor-aware QA synthesis. Our framework is generalizable and significantly reduces manual annotation effort.

\textbullet\ Our comprehensive evaluation of 30 SOTA video MLLMs offers key insights, facilitating in-depth discussions regarding their performance, strengths, and areas for enhancement.

 \textbullet\ We release our benchmark, including data, evaluation, and  synthesis codes at
\textit{\url{https://github.com/datajuicer/data-juicer/tree/HumanVBench}}
, to foster further evolution of future human-centric video analysis systems.

\section{Related Works}
\textbf{Video Benchmarks for MLLMs.} 
Recent years have witnessed the rapid development of benchmarks for evaluating MLLMs, particularly in the video domain. 
Representative efforts \citep{li2024mvbenchcomprehensivemultimodalvideo,fu2025video,zhang2025lvbench,zhou2025mlvu} such as Video-MME primarily target general video understanding abilities like temporal reasoning and broad scene comprehension. Although these include some human-centric content, their evaluation is often dispersed and lacks the structured, fine-grained focus necessary for deep diagnostics of human-centric perception.

To better assess human understanding, benchmarks often focus on observable activities like action recognition, pose estimation, or motion analysis~\citep{xu2016msr,tang2023humanbenchgeneralhumancentricperception,hong2025motionbench}, but typically neglect affective states, person recognition, or multimodal synchronization. In video emotion understanding, emotion-centric datasets~\citep{busso2008iemocap, poria2019meld, ren2024veatic} like VEATIC mostly rely on discrete emotion classification with fixed emotion categories, lacking multi-dimensional tasks such as emotion dynamics or cross-person intensity comparison. Broader MLLM benchmarks with emotion tasks often lack clear granularity and rigorous evaluation protocols. At a higher cognitive level, benchmarks~\citep{lei2018tvqa,lei2020tvqa, zadeh2019social} like Social-IQ target narrative reasoning, intention inference, and social intelligence, intertwining perception with higher-order reasoning and thus not isolating foundational perceptual competence.

In contrast, HumanVBench systematically targets the foundational perceptual layer of human-centric video understanding. It offers a comprehensive suite of 16 fine-grained tasks spanning person recognition, behavior and emotion perception, and audio-visual alignment, covering around 2.5k carefully synthesized multi-choice questions. This structured and multimodal evaluation serves as a prerequisite for higher-level cognitive reasoning. Quantitative comparisons with existing benchmarks are in Appendix~\ref{append:comparision_with_benchmark}.\\

\noindent\textbf{Benchmark Construction and QA Synthesis.} 
Methods for constructing QA pairs in existing video benchmarks fall into three categories: Some use fully manual annotation, with human-written questions and answers~\citep{wu2024longvideobench,fu2025video,zhang2025lvbench}.
 Others repurpose existing datasets, as in ActivityNet-QA~\citep{yu2019activitynet} built from ActivityNet~\citep{chen2019activitynet}, or HumanBench~\citep{tang2023humanbenchgeneralhumancentricperception}, which aggregates multiple public datasets for tasks like pose estimation and attribute recognition. Some benchmarks also incorporate automated QA generation, either by using commercial LLMs to produce QA pairs directly~\citep{li2024herm}, or by synthesizing reasoning and temporal questions from existing annotations like captions, temporal labels, or scene graphs~\citep{li2024mvbenchcomprehensivemultimodalvideo,yu2023anetqa,swetha2025timelogic}. However, these approaches typically rely on extensive manual or historical annotations and struggle to scale to in-the-wild video data from scratch. Moreover, the automated generation of challenging distractors remains underexplored. Existing methods—often based on general semantic similarity~\citep{stasaski2017multiple,kumar2023novel,qu2024unsupervised,feng2024exploring} or a single fine-tuned LLM for distractor generation~\citep{offerijns2020better,byun2025d}, lack mechanisms to systematically capture models’ actual inference weaknesses in the complex video domain.

To address these limitations, our method introduces a scalable, model-driven framework that constructs benchmarks directly from raw videos. This framework leverages a multi-operator approach for dense, fine-grained annotation and employs an MLLM ensemble to synthesize challenging and discriminative evaluations.

\section{The Proposed \ours}
\label{main:method}

\subsection{Task Design and Definition}
Human observers naturally focus on individuals in videos, attending to their appearance, emotions, and behaviors.
This intrinsic focus underpins our design of 16 fine-grained, human-centric tasks, aimed at evaluating MLLMs' ability to mimic human-like perception in video analysis (Figure \ref{fig:HumanVBench_fig}). 
Each task definitions and examples are provided in Appendix \ref{append:task_example}.
The tasks can be grouped into two categories based on content observability: \textit{Inner Emotion} and \textit{Outer Manifestation}.

Inner emotions are often less observable in real-world videos, requiring abstract skills to interpret subtle facial expressions, body language, and verbal cues. The tasks in this category, which we term \textbf{Emotion Perception}, evaluate MLLMs’ ability to detect and interpret these emotional nuances. Specifically, this category includes: 
\textit{Emotion Recognition} (ER) identifies the most fitting emotional description for an individual.
\textit{Emotion Temporal Analysis} (ETA) tracks emotional changes of a person over time.
\textit{Attitude Recognition} (AT) assesses the inferred attitude (positive, negative, or neutral) of individuals toward specific events or entities.
\textit{Emotion Intensity Comparison} (EIC) differentiates and quantifies the emotional intensity of various individuals.

In contrast to inner emotions, outer manifestations deal with more tangible aspects such as identifying individual(s), causality reasoning, and synchronization of video elements like speech or singing. Guided by these aspects of human observation, we formulated three task categories:

\textbf{Person Recognition} evaluates the model’s ability to identify individuals in complex scenes, similar to ``person-finding'':
\textit{Text-to-Human} (T2H) identifies a person based on a textual description.
\textit{Human-to-Text} (H2T) assesses the accuracy of a textual explanation attributed to a target person.
\textit{Human Counting} (HC) detects, tracks, and counts distinct individuals.
\textit{Appearance Time Detection} (ATD) identifies the specific time a specified individual appears.

\textbf{Behavior Analysis} scrutinizes a model's ability to understand and analyze individual behaviors, with four tasks:
\textit{Behavior Temporal Analysis} (BTA) tracks changes in a specified individual’s behaviors over time.
\textit{Behavior Causality Analysis} (BCA) examines causal reasoning of behaviors in the video context.
\textit{Action at Specified Time} (AST) identifies the precise actions occurring at a given time.
\textit{Time of Specific Action} (TSA) detects the exact moments when specific actions occur.

\textbf{Speech-Visual Alignment} is a critical cross-modal category designed to test capabilities like speaker identification and lip-syncing. This category includes:
\textit{Audio-Visual Speaker Matching} (AVSM) correlates audio features to identify individuals and their visual appearance (gender, age).
\textit{Active Speaker Detection} (ASD) identifies the individual currently speaking by integrating visual and audio inputs.
\textit{Audio-Visual Alignment Detection} (AVAD) detects synchronization points and coherence between lip movements and audio.
\textit{Speech Content Matching} (SCM) analyses spoken content against text, evaluating transcription or lip-reading capabilities.

Together, these tasks offer a comprehensive assessment of MLLMs' emotion, identity, behavior, and speech-visual alignment understanding, driving towards human-like video comprehension.

\begin{figure*}[t!]
\begin{center}
\centerline{\includegraphics[width=\linewidth]{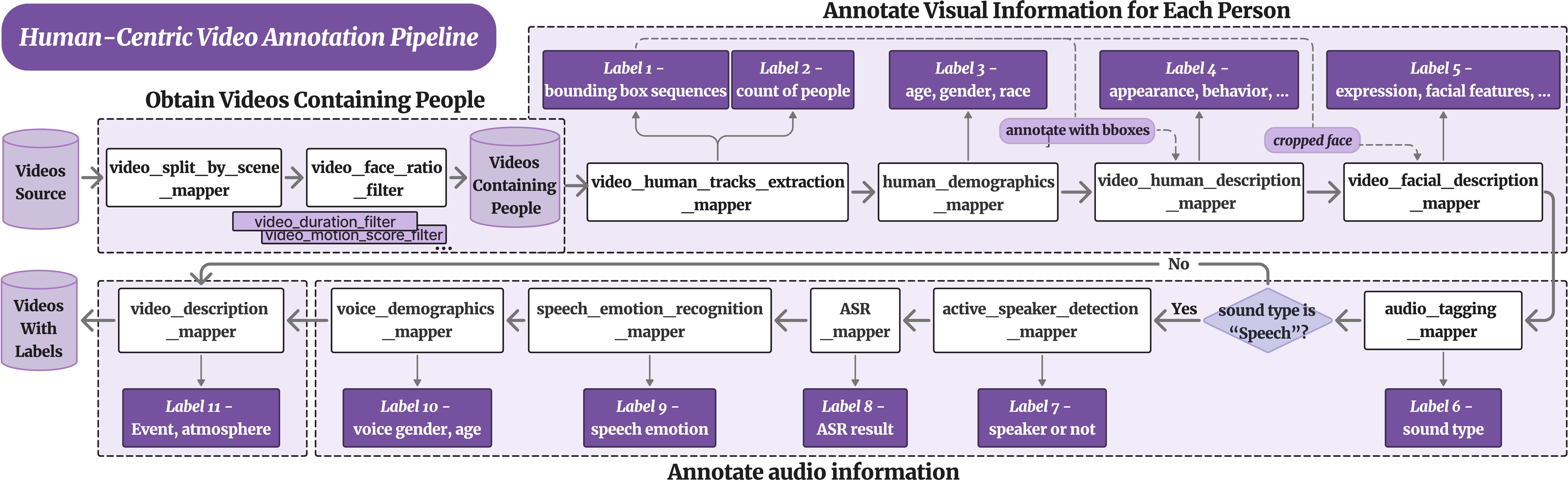}}
\vspace{-0.5em}
\caption{The Human-Centric Video Annotation Pipeline involves obtaining videos featuring people and annotating both visual and auditory information as well as overall event atmospheres.}
\label{fig:Annotation pipeline}
\end{center}
\vspace{-2em}
\end{figure*}

\begin{figure*}[t!]
\begin{center}
\centerline{\includegraphics[width=\linewidth]{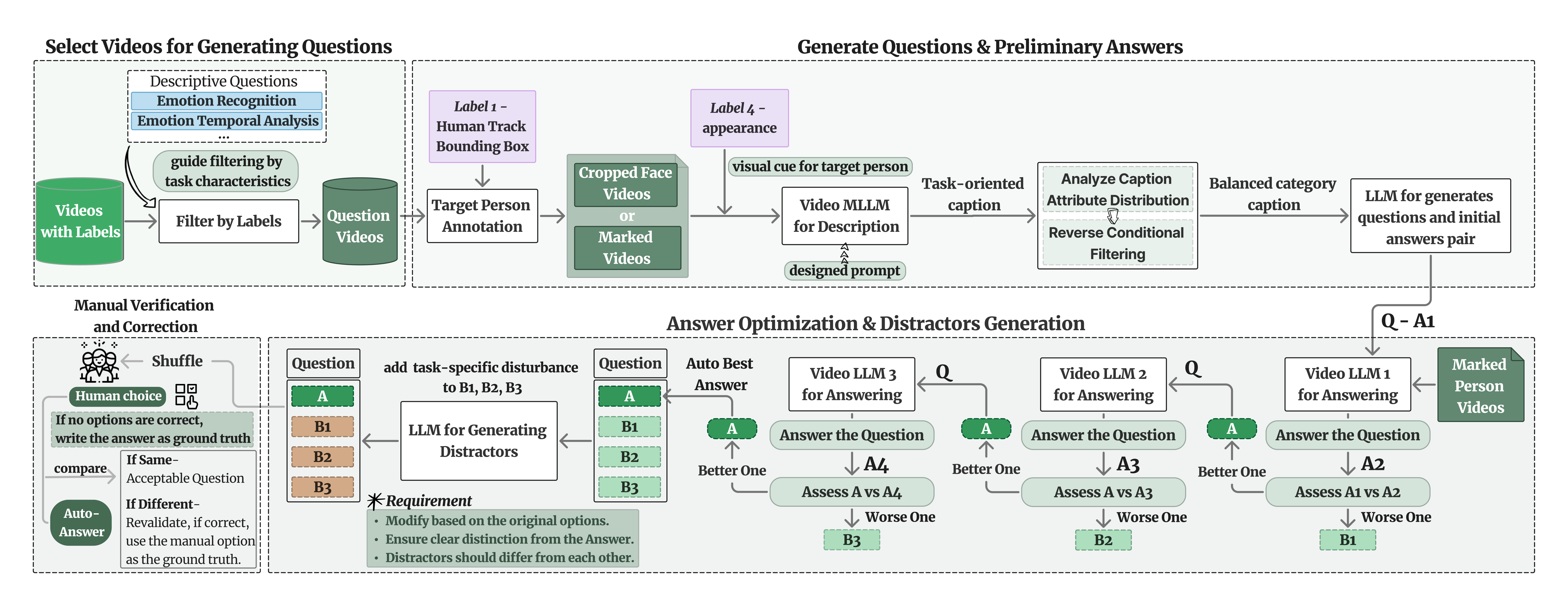}}
\caption{The Distractor-Included QA Synthesis Pipeline facilitates four steps: selecting ``question videos'', generating Questions, ensemble-based options generation with model-derived errors as distractors, and manually verifying  multiple-choice questions.}
\label{fig:MCQ Generation Pipeline}
\end{center}
\vspace{-2.5em}
\end{figure*}

\subsection{Human-Centric Video Annotation Pipeline}\label{section:Annotation Pipeline}
Creating task-specific questions for the aforementioned tasks hinges on extensive human-centric annotations within videos. Our video annotation pipeline, illustrated in Figure \ref{fig:Annotation pipeline}, emphasizes multi-modal, granular annotation pathways, supported by the Data-Juicer framework \citep{chen2024data,djsandbox}. While benefiting from existing operators, we've also developed innovative operators to enhance the open-source community. Detailed annotations and examples are in the Appendix \ref{append:annotation_pipeline}.

\textbf{Collecting Videos Containing People.}
We primarily sourced copyright-free videos from \href{https://www.pexels.com/}{Pexels}, which offers diverse in-the-wild footage under varied scenes and capture conditions. Movies from the MF2 benchmark~\citep{zaranis2025moviefactsfibsmf2}, licensed under Public Domain 1.0, were also included to further increase content diversity.
Importantly, we used only the raw visual data without any pre-existing captions or metadata. All annotations were generated internally by our pipelines, thereby preventing risks of data leakage.
We begin by splitting each video into scenes with {\small \texttt{video\_split\_by\_scene\_mapper}} to facilitate subsequent accurate human tracking. The resulting clips are then filtered by attributes such as duration and optical flow, after which {\small\texttt{video\_face\_ratio\_filter}} is applied to ensure that faces are visible in most frames. This guarantees that the selected videos are primarily human-centric and ready for further annotation.

\textbf{Human-Centric Video Annotation Pipeline.}    
We design a set of operators to extract and enrich human-centered video information. First, {\small\texttt{video\_human\_tracks\_extraction\_mapper}} links detected faces and bodies across consecutive frames using overlap thresholds, producing reliable person tracks and an approximate count of individuals in a shot. These tracks support subsequent tasks such as person highlighting and appearance/action descriptions.

From the tracks, {\small\texttt{human\_demographics\_mapper}} extracts face crops and applies attribute models to infer demographic labels (e.g., age, gender, race). Flexible bounding box crops (full-body or face-only) further allow generating individual-focused clips. Based on this, {\small\texttt{video\_human\_description\_mapper}} captures appearance and posture using an MLLM, while {\small\texttt{video\_facial\_description\_mapper}} focuses on facial expressions and changes. This ensures descriptions are unaffected by background or other people, while preserving facial detail.

For audio, {\small\texttt{audio\_tagging\_mapper}} first classifies sound types. If speech is present, additional operators are invoked: {\small\texttt{active\_speaker\_detection\_mapper}} fuses audio-visual cues to localize speakers, {\small\texttt{asr\_mapper}} transcribes speech, {\small\texttt{speech\_emotion\_recognition\_mapper}} detects emotions, and {\small\texttt{voice\_demographics\_mapper}} profiles vocal attributes (e.g., gender, age).

Finally, {\small\texttt{video\_description\_mapper}} summarizes overall atmosphere and narrative using MLLMs. These multimodal annotations enable automated task construction. For example, in the \textit{Audio-Visual Speaker Matching} task, videos are selected where exactly one individual’s visual demographics (\textit{Label-3}) match the audio demographics (\textit{Label-10}), while other individuals’ features mismatch. 
The matched individual serves as ground truth, with IDs assigned via \textit{Label-1} tracking. Implementation details for each operator and full task construction appear in Appendices \ref{append:annotation_pipeline} and \ref{append:annotation_details_all_task}.

\vspace{-0.2em}
\subsection{Distractor-Included QA Generation Pipeline}
For tasks with exact answers (e.g., restricted categories, numbers, or letters), questions, correct answers, and distractors are constructed using tailored templates and annotations from Section \ref{section:Annotation Pipeline}. For open-ended questions, we developed a pipeline to generate questions, answers, and distractors, 
applied to six tasks in \ours: \textit{Emotion Recognition, Emotion Temporal Analysis, Emotion Intensity Comparison, Human-to-Text, Behavior Temporal Analysis}, and \textit{Behavior Causality Analysis}. The pipeline (Figure~\ref{fig:MCQ Generation Pipeline}) consists of four stages:

\textbf{Selecting Videos.}
Videos are filtered by task criteria. For example, \textit{Behavior Temporal Analysis} requires longer clips, while \textit{Human-to-Text} requires at least two people.

\textbf{Question and Preliminary Answer Synthesis.}\label{Synthesizing Questions and Preliminary Answers} Firstly, target persons are localized using bounding box tracks. For facial emotion-focused tasks, we reconstruct the video with cropped face regions; for others, the target person is highlighted with a red bounding box to create ``marked videos''.
These marked videos are fed into a Video-MLLM with prompt designed for the specific task, so that the model produces targeted descriptions such as emotional states or behavior summaries. Since Video-MLLMs may not always focus on the highlighted individual, we incorporate a visual cue from \textit{Label-4} (Figure~\ref{fig:Annotation pipeline}) into the prompt. The generated descriptions are then analyzed to extract task-relevant attributes, and their distributions are adjusted to ensure balance (e.g., maintaining similar proportions of positive and negative emotions). Based on these captions, an LLM (GPT-4 in our workflow) is used to generate questions and preliminary answers.

\textbf{Ensemble Answer Selection and Distractors Generation.}
To mitigate errors from single-model answers, we adopt an ensemble strategy. Multiple MLLMs (Gemini~\citep{team2023gemini}, VideoLLaMA3~\citep{zhang2025videollama}, ShareGPT4Video~\citep{chen2024sharegpt4video}) produce candidate answers, which are ranked via preference voting. The top choice is taken as the correct answer, while the others are converted into distractors: if semantically distinct (valid answer component), they are retained; otherwise an LLM (GPT-4 in our workflow) introduces task-specific perturbations. Retaining these erroneous answers allows the distractors to reflect typical model mistakes, thereby ensuring plausibility and distractive power rather than being arbitrarily fabricated, which increases the overall difficulty of the questions.

\textbf{Manual Verification and Correction}\label{main:Manual Verification}
Given the annotation models’ limitations, systematic biases and errors may occur in the generated multiple-choice questions. To address this concern and ensure the benchmark’s reliability, we incorporate a human verification stage: all candidate options are shuffled, and annotators are instructed to select the correct one; if none of the options is fully accurate, they directly write the correct answer, which is then incorporated as the ground truth.

The Distractor-Included QA Synthesis Pipeline handles large-scale question and distractor generation, while humans serve as final arbiters of correctness. This substantially reduces manual workload—annotators only needed to rewrite the correct answer in about 25\% of cases, while in the remaining 75\% they confirmed one of the generated options (see Section \ref{Quantifying_Manual_Labor}). To further ensure reliability, we conducted an inter-annotator agreement (IAA) study on 240 randomly sampled questions from the final verified questions, where two independent annotators achieved a Cohen’s Kappa of 0.8833, demonstrating the reliability of our benchmark. Additionally, we also provide distractor quality evaluation in Appendix~\ref{append:distractor_evaluation}.

\subsection{Post-Processing}
To address the prevalent issue of answer leakage in multimodal evaluation datasets, we adopt the approach proposed in \citep{chen2024we}. Specifically, we test models without visual input and remove frequently correct QAs (around 6\%), ensuring random accuracy. This preserves visual relevance while mitigating leakage risks. In total, \ours contains 2475 problem instances; per-task counts are shown in Figure \ref{fig:HumanVBench_fig}, with additional statistics in Appendix~\ref{append:statistics}.

\begin{table*}[t]
\belowrulesep=0pt
\aboverulesep=0pt
\renewcommand{\arraystretch}{1.0} 
\scriptsize 
\setlength{\tabcolsep}{3pt} 
\setlength{\abovecaptionskip}{0.2cm}  
\setlength{\belowcaptionskip}{-0.2cm} 
\centering
\caption{Summary of performance on \ours. 
We report average accuracy (\%) across four main categories. Full results for all 30 models and 16 tasks are in Appendix \ref{append: Full_Results}.
The best overall results are shown in bold, while the best open-source model results are underlined ``-'' means the model recognizes its lack of required capabilities for the task and thus refuses to answer.
}
\label{tab:summary_results}
\begin{tabular}{l|c|c|c|c|>{\columncolor{gray!20}}c|cccc|c|>{\columncolor{gray!20}}c}
\toprule
\multirow{2}{*}{Model} & \multirow{2}{*}{Mod.} & \textbf{Emotion} & \textbf{Person Recog.} & \textbf{Behavior} & \textbf{Avg.}& \multicolumn{5}{c|}{\textbf{Speech-Visual Alignment}} & \textbf{Overall Avg.} \\
& & 4 tasks & 4 tasks & 4 tasks & 12 tasks & AVSM & ASD & AVAD & SCM & 4 tasks & 16 tasks \\
\hline
\multicolumn{2}{l|}{Random Guess} & 24.4 & 25.2 & 22.9 & 24.2 & 42.8 & 23.6 & 33.3 & 25.0 & 31.2 & 25.9\\
\hline
Qwen-VL3 (7B)& V & 43.2& 67.6& 54.3& \underline{55.0}& \underline{71.1}& \underline{65.8}& \underline{34.3}& 22.2& 48.3& \underline{53.4}\\
InternVL3.5 (8B)& V & \underline{47.5}& 57.4& 58.8& 54.6& 63.8& 63.2& 31.4& 23.1& 45.4& 52.3\\
VideoLLaMA3 (7B) & V & 39.7 & \underline{68.5}& \underline{55.8}& 54.7& 64.6 & 63.1& \underline{34.3}& 17.9 & 45.0 & 52.3\\
\hline
VideoLLaMA (7B) & V+A & 21.1 & 23.8 & 25.7 & 23.5 & 40.1 & 26.6 & 33.1 & 26.2 & 31.5 & 25.5\\
VideoLLaMA2.1 (7B)& V+A  & 36.5 & 36.4 & 43.6 & 38.8 & 44.0 & 31.6 & 32.1 & 23.7 & 32.9 & 37.4\\
Qwen2.5-Omni (7B)& V+A & 35.5 & 44.5 & 38.3 & 39.4 & \underline{71.1}& 48.4& 27.0& \underline{71.8}& \underline{54.6}& 43.2\\

\hline
GPT-4o & V & 33.6 & 50.9 & 62.1 & 48.9 & - & - & - & - & - & -\\
GPT-5 & V & 46.8 & 69.5 & 67.3 & 61.2 & - & - & - & - & - & -\\
Gemini-1.5-Pro & V+A & 50.9 & 71.4 & 60.7 & 61.0 & 90.1& 76.8& 66.4& 84.6& 79.5& 65.6\\
Gemini-2.5-flash & V+A & \textbf{53.0}& 75.7 & 64.9 & 64.5 & \textbf{98.7}& \textbf{80.6}& 57.7 & \textbf{99.1}& 84.0 & 64.5\\
Gemini-2.5-Pro & V+A & 52.9 & \textbf{83.5}& \textbf{70.7}& \textbf{69.0}& 96.7 & 78.7 & \textbf{72.3}& 98.3 & \textbf{86.5}& \textbf{73.4}\\
\hline
\multicolumn{2}{l|}{Human (Graduate)} & 84.6 & 88.5 & 87.0 & 86.7 & 96.0 & 96.1 & 87.0 & 98.3 & 94.4 & 88.6\\
\bottomrule
\end{tabular}%
\end{table*}

\section{Evaluation and Insights}
\subsection{Experimental Settings}

We evaluate 30 of the most popular video MLLMs, including visual-only MLLMs from the Qwen-VL series,
InternVL series, and VideoLLaMA3, as well as audio-visual MLLMs like Qwen2.5-Omni \citep{xu2025qwen25omnitechnicalreport} and Video-LLaMA series \citep{zhang2023video, cheng2024videollama}. We also evaluated commercial models from the GPT-4o, GPT-5, and Gemini series. All tasks were cast as multiple-choice questions (N-choose-1, with variable N), with human and random guesses as baselines. For more implementation details and the full results of all 16 tasks on 30 MLLMs , please refer to Appendix \ref{append:eval_detail} and \ref{append: Full_Results}.

\subsection{Comprehensive Evaluation of MLLMs}
Table \ref{tab:summary_results} summarizes the performance evaluation results for our benchmark, revealing several key insights from analyzing vision-only tasks (emotion perception, person recognition, and behavior analysis tasks) and cross-modal tasks (speech-visual alignment tasks). 
\subsubsection{Performance in Vision-Only Tasks}
\textbf{Open-Source Video-MLLMs:} Overall, current models still exhibit a clear gap from human-level performance, especially in Emotion Perception tasks, underscoring their limited ability to capture emotional nuances. Nevertheless, among open-source models, Qwen3-VL leads in video human understanding, and achieves a modest 55.0\% mean accuracy across 12 vision-only tasks. 
Although it still lags behind human performance, it has surpassed GPT-4o and approaches other commercial counterparts, highlighting the immense potential of open-source models.

\noindent \textbf{Proprietary MLLMs:} 
Overall, with the exception of GPT-4o, proprietary models substantially outperform open-source ones across tasks, though their capability in emotion perception remains relatively weak, highlighting the need for better nuanced emotion understanding; notably, Gemini-2.5-Pro stands far ahead, achieving near-human proficiency in person recognition, whereas GPT-4o performs unexpectedly poorly; its subpar performance in the Emotion Understanding category and in specific Person Recognition tasks (T2H and ATD) results in its average performance being lower than that of several leading open-source models.

\subsubsection{Performance in Speech-Visual Alignment Tasks}
\textbf{Lip-Reading Ability of Visual-Only MLLMs} For AVSM and ASD tasks, many videos include dubbing or multi-speaker conversations with single-speaker audio, making lip movements alone insufficient for accurate responses. Therefore, when answering these two tasks, visual-only models essentially degrade to speech action recognition but can still get some questions correct, with QwenVL3 outperforming other models. However, for the AVAD and SCM task, almost all models perform at a random level, indicating that current models lack precise lip-reading (lip translation) ability. Future models could focus on enhancing lip-reading capabilities.

\noindent \textbf{Audio-Video MLLMs:} 
Although most open-source MLLMs can process both audio and visual inputs, they—except for Qwen2.5-Omni—perform near-randomly on Speech-Visual Alignment tasks, particularly AVSM and ASD, where they even lag behind visual-only models.
This deficiency may stem from two key factors: a limited ability to visually interpret intricate lip movements (as further detailed in Appendix Section \ref{append:ModalityAblation}) and a general lack of a suitable temporal processing mechanism to facilitate fine-grained speech-to-lip alignment, further exacerbated by scarce datasets providing audio-visual lexical mappings. Consequently, while dedicated ASR models can effectively handle SCM, most open-source video-MLLMs struggle acutely with correlating speech to lip movements. In contrast, the Gemini series shows exceptional cross-modal alignment, performing robustly even on AVAD, a task no other models can handle.

\begin{figure*}[t!] 
\centering
\includegraphics[width=\linewidth]{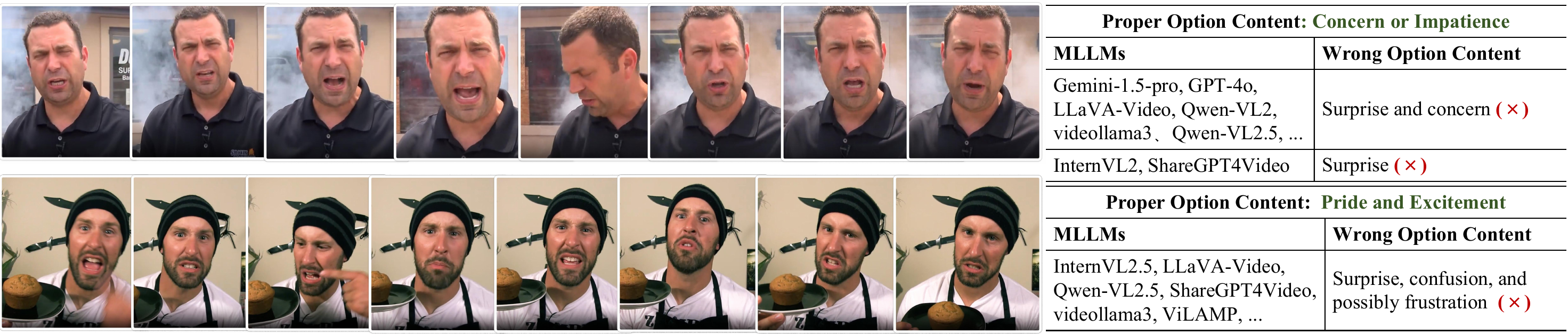}
\caption{Two examples of 8-frame speaker videos sampled evenly for the emotion recognition task, with responses from various MLLMs.}
\label{fig:speaker ER}
\end{figure*}
\subsection{Dissecting Fine-grained Insights}
\subsubsection{Discussion on Speaker Emotion Recognition}\label{Challenges in Speaker Emotion Recognition}

\begin{table*}[t] 
\scriptsize 
\renewcommand{\arraystretch}{1.1}
\centering
\vspace{-0.5em}

\begin{subtable}{0.28\textwidth}
    \centering
    \setlength{\tabcolsep}{4pt}
    
    \caption{\textit{Emotion Recognition} Accuracy for the full dataset and the subset where target individuals are in the speaking state.}
    \label{table:Speaker_Emotion_Recognition}
    
    \begin{tabular}{l|c|c}
    \hline
    \multirow{2}{*}{\textbf{Models}} & \multicolumn{2}{c}{\textbf{ER Accuracy}} \\
    \cline{2-3} 
    & Full & Speaking Subset \\
    
    \hline
    CogVLM2 & 38.4 & 34.7\textcolor{DarkGreen}{($\downarrow$3.7)} \\
    VideoLLaMA3 & 34.6 & 30.3\textcolor{DarkGreen}{($\downarrow$4.3)} \\
    LLaVAOneVision & 36.9 & 34.7\textcolor{DarkGreen}{($\downarrow$2.2)} \\
    InternVL3.5 & 43.7 & 40.9\textcolor{DarkGreen}{($\downarrow$2.8)} \\
    InternVideo & 35.7 & 31.0\textcolor{DarkGreen}{($\downarrow$4.7)} \\
    Qwen-VL2.5 & 43.0 & 41.0\textcolor{DarkGreen}{($\downarrow$2.0)} \\
    ShareGPT4Video & 35.0 & 31.6\textcolor{DarkGreen}{($\downarrow$3.4)} \\
    ChatBridge & 28.9 & 24.1\textcolor{DarkGreen}{($\downarrow$4.8)} \\
    Qwen2.5-Omni & 42.2 & 40.3\textcolor{DarkGreen}{($\downarrow$1.9)} \\
    \hline
    Average & 36.9 & 33.5\textcolor{DarkGreen}{($\downarrow$3.3)} \\
    \hline
    \end{tabular}
\end{subtable}
\hfill 
\begin{subtable}{0.33\textwidth}
    \centering
    \setlength{\tabcolsep}{3pt}

    \caption{Timestamp integration effect on Video-MLLMs in \textit{Appearance Time Detection} , \textit{Action at Specific Time} and \textit{Time of Specific Action}.}
    \label{table:Add_ts}

    \begin{tabular}{l|c|c|c}
    \hline
    \multirow{2}{*}{\textbf{Models}} & \multicolumn{3}{c}{\textbf{Time-Specific Tasks}} \\
    \cline{2-4} %
    & ATD & AST & TSA \\
    \hline
    Chat-UniVi & 22.1 \textcolor{red}{($\uparrow$2)} & 40 \textcolor{red}{($\uparrow$26)} & 14.1 \textcolor{red}{($\uparrow$7)} \\
    CogVLM2 & 40.8 \textcolor{red}{($\uparrow$1)} & 35 \textcolor{DarkGreen}{($\downarrow$1)} & 10.7 \textcolor{DarkGreen}{($\downarrow$3)} \\
    VILA & 38.3 \textcolor{red}{($\uparrow$18)} & 33 (--) & 48 \textcolor{DarkGreen}{($\downarrow$3)} \\
    Video-LLaVA & 36.7 \textcolor{red}{($\uparrow$5)} & 25 \textcolor{DarkGreen}{($\downarrow$2)} & 45.8 \textcolor{red}{($\uparrow$5)} \\
    LLaVAOneVision & 10 (--) & 41 \textcolor{red}{($\uparrow$4)} & 47.5 (--) \\
    InternVL2 & 31.7 \textcolor{red}{($\uparrow$11)} & 45 \textcolor{red}{($\uparrow$7)} & 52.5 \textcolor{red}{($\uparrow$19)} \\
    Video-LLaMA & 24.2 \textcolor{red}{($\uparrow$4)} & 25 \textcolor{red}{($\uparrow$7)} & 9 (--) \\
    Video-LLaMA2.1 & 24.2 \textcolor{red}{($\uparrow$7)} & 25 \textcolor{DarkGreen}{($\downarrow$2)} & 29.4 \textcolor{DarkGreen}{($\downarrow$4)} \\
    ChatBridge & 17.5 \textcolor{red}{($\uparrow$9)} & 40 \textcolor{DarkGreen}{($\downarrow$1)} & 14.1 \textcolor{DarkGreen}{($\downarrow$3)} \\
    \hline
    Average & 26.8 \textcolor{red}{($\uparrow$6)} & 34.3 \textcolor{red}{($\uparrow$4)} & 30.1 \textcolor{red}{($\uparrow$2)} \\
    \hline
    \end{tabular}
\end{subtable}
\hfill %
\begin{subtable}{0.34\textwidth}
    \centering
    \setlength{\tabcolsep}{2pt}

    \caption{Effectiveness of the Distractor-Included QA Synthesis Pipeline in generating six types of descriptive multiple-choice questions. The Efficiency Gain is calculated as 1 / (1 - No Edit Rate).}
    \label{tab:qa_effectiveness}

    \begin{tabular}{lcc}
    \toprule
    \textbf{Question Type} & \textbf{No Edit} & \textbf{Efficiency Gain} \\
    \midrule
    \textit{Behavior Temporal Analysis} & 68\% & 3.1 \\
    \textit{Emotion Intensity Comparison} & 63\% & 2.7 \\
    \textit{Emotion Recognition} & 81\% & 5.3 \\
    \textit{Behavior Causality Analysis} & 74\% & 3.8 \\
    \textit{Emotion Temporal Analysis} & 83\% & 5.9 \\
    \textit{Human-to-Text} & 65\% & 2.9 \\
    \midrule
    Average & 72.3\% & 3.6 \\
    \bottomrule
    \end{tabular}
\end{subtable}
\vspace{-0.5em}
\end{table*}

A detailed analysis reveals that models frequently misattribute speaker emotions as ``surprise'' or ``shock''.  This issue primarily stems from frame sampling processes, where frames depicting ``mouth-opening'' motions are misclassified. Figure \ref{fig:speaker ER} illustrates this phenomenon using the commonly adopted eight-frame fixed sampling approach \citep{lin-etal-2024-video, zhang2024llavahddivinghighresolutionlarge}, which introduces temporal noise and leads to erroneous emotional judgments. The speaker-centric evaluations presented in Table \ref{table:Speaker_Emotion_Recognition} further confirm a consistent decline in accuracy across all models, underscoring the inherent challenges of this task for video MLLMs.

\subsubsection{Timestamps Impact on Time-Specific Tasks}
Our findings show that many open-source video-MLLMs struggle with time-specific tasks due to their inability to correlate video events with a timeline (Appendix Table \ref{append: Full_Results}). While top performers like VideoLLaMA3 and LLaVA-Video natively incorporate textual timestamps, we explored this capability by adding timestamps to models without native support. As shown in Table \ref{table:Add_ts}, this intervention generally improves temporal inference accuracy, though its effectiveness varies due to models’ limited exposure to timestamp data. Despite these gains, a significant gap remains compared to MLLMs that natively incorporate time-event alignment, underscoring the necessity of integrating temporal reasoning mechanisms from the outset of model design.

\subsection{Quantifying the Reduction of Manual Labor in Question Generation}\label{Quantifying_Manual_Labor}
We report the distribution of manual work required for the 6 descriptive multiple-choice questions generated by the pipeline in Table \ref{tab:qa_effectiveness}. 
On average, 72\% of the generated questions contain the correct answer, requiring no revision, with humans only needing to select the correct answer. 
The remaining 28\% lack a correct answer and require a simple manual input, typically one sentence, which is much faster than creating questions from scratch.
Notably, the overall proportion of questions containing the correct answer exceeds the best model’s accuracy in the final evaluation (e.g., the “no-edit” rate for ER is 81\%, while the best model accuracy is 49.4\%), as options are drawn from multiple models, increasing correct answer coverage and reflecting the benefit of ensemble-based design. Besides, incorrect model responses are repurposed as plausible distractors, enhancing question difficulty. For other rule-based non-descriptive tasks, over 70\% of the questions can be directly adopted. 
Overall, while human oversight is still needed, our partial automation significantly boosts efficiency, allowing a person using the pipeline to complete 3 times more questions within the same timeframe as manual creation from scratch, substantially reducing manual workload.

\begin{figure*}[h!]
    \centering
    \includegraphics[width=\linewidth]{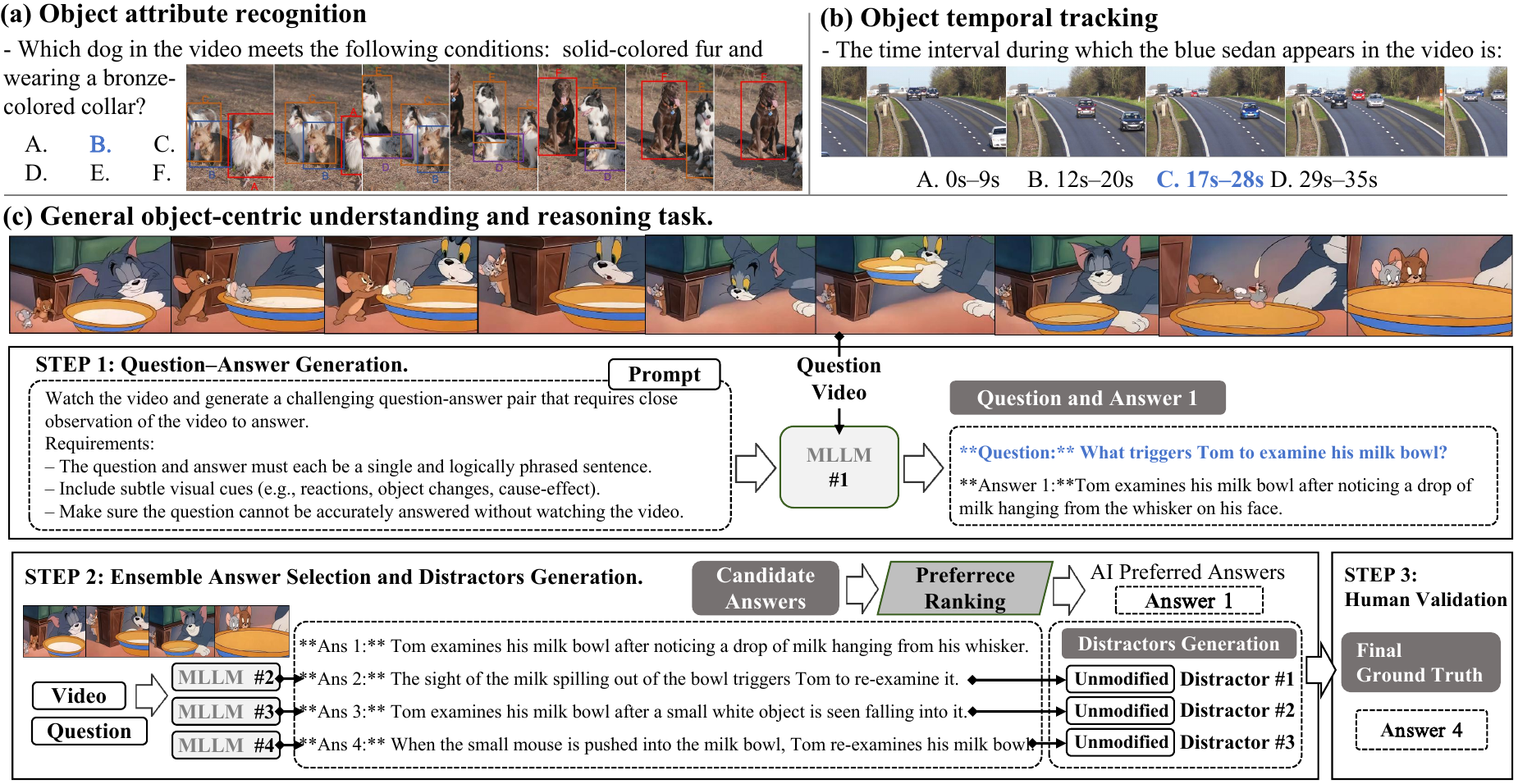}
    \vspace{-1.5em}
    \caption{Demonstration of our pipeline's generalizability. (a) Object attribute recognition for pets. (b) Object temporal tracking for cars. (c) General causal reasoning QA generated for an object-centric event, with the MLLMs used in this case being Gemini-1.5-pro, VideoLLaMA3, LLaVA-Video, and gpt-4o. This highlights the adaptability of our framework.}
    \label{fig:pipeline_generalizability}
    \vspace{-0.5em}
\end{figure*}

\subsection{Generalizability of the Synthesis Pipelines Beyond Human-Centric Tasks} \label{exp:pipeline_extension}
A key contribution of our work is the general-purpose nature of our automated pipelines, which extends far beyond the human-centric domain. Its modular architecture allows for easy adaptation to new domains by simply swapping initial detectors. We demonstrate this by generating diagnostic questions for non-human entities, such as fine-grained attribute recognition for pets and temporal tracking for vehicles, with examples shown in Figure \ref{fig:pipeline_generalizability}.

For object attribute recognition (e.g., a specific dog; Figure~\ref{fig:pipeline_generalizability} (a)), we adapt our pipeline by: \textit{(i)} replacing the human detector with a dog detector (YOLOv11) to extract instance tracks (\textit{Label-1}); \textit{(ii)} using an MLLM to annotate attributes (e.g., fur pattern, accessories) from cropped tracks; \textit{(iii)} selecting a uniquely described dog as the ground truth; and \textit{(iv)} rendering bounding boxes from \textit{Label-1} as candidates. 
Similarly, for object temporal tracking (e.g., a car’s appearance interval, Figure~\ref{fig:pipeline_generalizability} (b)), we: \textit{(i)} replace the human detector with a vehicle detector for tracking; \textit{(ii)} annotate attributes (e.g., color, type) for each track; \textit{(iii)} select a unique car as ground truth; and \textit{(iv)} use track frames to determine the appearance interval as ground truth.
These procedures parallel our \textit{Text-to-Human} and \textit{Appearance Time Detection} tasks, proving the framework's broad applicability to object-centric attribute recognition and temporal tracking. 

Furthermore, our Distractor-Included QA Generation Pipeline is inherently transferable, enabling the synthesis of diverse descriptive questions and answer choices beyond human-centric scenarios, as exemplified by the general causal reasoning question in Figure \ref{fig:pipeline_generalizability} (c). This transferability underscores the broader value of our methodology as a scalable solution for accelerating the creation of high-quality MLLM evaluations across diverse domains.

\section{Conclusion}
\label{sec:conclusion}
We present \ours to address the pressing need for improved assessment of human-centric video understanding in MLLMs. 
By incorporating extensive evaluation dimensions through 16 fine-grained tasks, \ours provides a systemic view into both successes and critical shortcomings of video MLLMs, particularly in emotion perception and speech-visual alignment. Experimental findings across over twenty leading video MLLMs illustrate that while proprietary models approach human accuracy in some tasks, substantial advancements are required, particularly in cross-modal domains where alignment between speech and visual elements proves challenging. 

Beyond the benchmark itself, the core contribution of this work lies in its extensibility and efficient methodology. Our framework transforms model-generated errors into high-quality distractors and shifts the human role from creating tasks from scratch to efficiently validating and correcting them, thereby multiplying the benchmark's production efficiency.
By open-sourcing the benchmark and underlying methodologies, we hope \ours can foster collaborative efforts aimed at advancing the frontiers of human-like video understanding capabilities in MLLMs.

\section*{Acknowledgments}
This work was supported in part by the New Generation Artificial Intelligence-National Science and Technology Major Project (2025ZD0123003), the National Natural Science Foundation of China Enterprise Innovation and Development Joint Fund (Artificial Intelligence Field) Key Support Projects (U25B2072), and The Major Key Project of PCL (Grant No. PCL2025A17).

{
    \small
    \bibliographystyle{ieeenat_fullname}
    \bibliography{main}
}

\appendix
\clearpage
\setcounter{page}{1}
\maketitlesupplementary

\newpage

\section*{Appendix Contents}
\begin{enumerate}[label=\Alph*.]
    \item \textbf{The Use of Large Language Models} \dotfill \pageref{append:LLM use}
    \item \textbf{Comparison with Existing Video Benchmarks} \dotfill \pageref{append:comparision_with_benchmark}
    \item \textbf{More Statistics of \ours} \dotfill \pageref{append:statistics}
    \item \textbf{Full Results for 24 Models on 16 Tasks} \dotfill \pageref{append: Full_Results}
    \item \textbf{Distractor Quality Analysis} \dotfill \pageref{append:distractor_evaluation}
    \item \textbf{Modality Ablation in VideoLLaMA2} \dotfill \pageref{append:ModalityAblation}
    \item \textbf{Impact of Frame Sampling} \dotfill \pageref{append:rebuttal_frame}
    \item \textbf{Impact of Human Density on Model Performance} \dotfill \pageref{append:rebuttal_human_density}
    \item \textbf{Model Evaluation Implementation} \dotfill \pageref{append:eval_detail}
    \item \textbf{Definitions and Examples for Each Task} \dotfill \pageref{append:task_example}
    \item \textbf{Annotations Details in Human-Centric Annotation Pipeline} \dotfill \pageref{append:annotation_pipeline}
    \item \textbf{Complete Construction Details of All Tasks} \dotfill \pageref{append:annotation_details_all_task}
\end{enumerate}


\section{The Use of Large Language Models}
\label{append:LLM use}
The core research idea, including the conceptualization of HUMANVBENCH and its synthesis framework, originated entirely from the authors. Multimodal Large Language Models (MLLMs) and Large Language Models (LLMs) were then integrally utilized as a methodological component to efficiently automate data annotation and challenging question construction for the benchmark, a key focus of our research. Details of MLLM utilization, including the models, prompts, and annotation pipeline, are provided in the methodology section of the main paper (Section \ref{main:method}) and further elaborated in Appendix~\ref{append:annotation_pipeline} and Appendix~\ref{append:annotation_details_all_task}. Additionally, LLMs assisted with language refinement during manuscript preparation. The authors assume full responsibility for all content, including research ideas, methodologies, and findings.

\section{Comparison with Existing Video Benchmarks}
\label{append:comparision_with_benchmark}
\begin{table*}[t]
\scriptsize
\centering
\setlength{\tabcolsep}{3pt}  
\renewcommand{\arraystretch}{1.15}
\caption{Comparison of \ours with existing video-based MLLM benchmarks.}
\label{table: Comparison_benchmarks}
\begin{tabular}{p{1.4cm} p{2.2cm} p{0.7cm} p{0.7cm} p{1.5cm} 
                p{0.8cm} p{1.8cm} p{1.5cm} 
                p{0.9cm} p{1.5cm} p{0.9cm} p{1.0cm}}
\toprule
\multirow{2}{*}{\textbf{Benchmark}} & 
\multirow{2}{*}{\textbf{Feature}} &
\multirow{2}{*}{\textbf{Video}} &
\multirow{2}{*}{\textbf{\#QA}} &
\multirow{2}{*}{\makecell{\textbf{Human-}\\\textbf{centric QA}}} &
\multirow{2}{*}{\makecell{\textbf{Video}\\\textbf{Length}}} &
\multirow{2}{*}{\makecell{\textbf{Annotation}\\\textbf{Source}}} &
\multirow{2}{*}{\makecell{\textbf{QA}\\\textbf{Generation}}} & 
\multicolumn{4}{c}{\textbf{Evaluation Content}} \\
\cmidrule(lr){9-12}
& & & & & & & &
\textbf{Emotion} & \textbf{Person Recog.} & \textbf{Behavior} & \textbf{Lip-sync} \\
\midrule

\rowcolor{gray!10} Video-MME & General Video \newline Comprehension & 900 & 2,700 & 1,084 \newline (40.1\%) & $<$1h & -- & Fully human \newline annotated & 
33 cases \newline (1\%) & \ding{51} & \ding{51} & \ding{55} \\

MVBench & General Video \newline Comprehension & 3,259 & 3,600 & 1,684 \newline (46.7\%) & $<$35s & Existing Captions & LLM-based & 
33 cases \newline (0.9\%) & 121 cases \newline (3\%) & \ding{51} & \ding{55} \\

\rowcolor{gray!10} ActivityNet-QA (Test Set) & General Video \newline Comprehension & -- & 8,000 & 1,050 \newline (13.1\%) & $\sim$3min & -- & Fully human \newline annotated & 
\ding{55} & \ding{51} & \ding{51} & \ding{55} \\

TLQA & Temporal Logic \newline Reasoning & -- & 160k & -- & $<$5min & Existing temporal labels & Automatic logic templates & 
-- & -- & -- & \ding{55} \\

\rowcolor{gray!10} ANetQA & Compositional \newline Reasoning & -- & 13.4M & -- & $\sim$3min & Manually annotated scene graphs & Automatic template-based & 
-- & -- & -- & \ding{55} \\

VEATIC & Continuous Emotion (Valence/Arousal) & 124 & -- & 100\% & $<$2m37s & Real-time human labeling & - & 
\ding{51} & -- & -- & \ding{55} \\

\rowcolor{gray!10} TVQA (Test Set) & TV Plot \newline Understanding & 1,809 & 7,623 & 100\% & $<$90s & -- & Fully human annotated & 
-- & -- & -- & \ding{55} \\

Social-IQ & Social Situations Reasoning & 1,250 & 7,500 & 100\% & $<$30s & -- & Fully human annotated & 
-- & -- & -- & \ding{55} \\

\rowcolor{gray!10}
HumanVBench & Foundational \newline Human Perception & 2,182 & 2,475 & 100\% & $<$30s & Multi-operator auto-annotation & Rule-based \& LLM ensemble & 
\ding{51} & \ding{51} & \ding{51} & \ding{51} \\

\bottomrule
\end{tabular}
\end{table*}

Table \ref{table: Comparison_benchmarks} compares existing video benchmarks for MLLMs, highlighting their limited focus on foundational human perception tasks such as emotion understanding and audio-visual alignment. Moreover, most rely on manual annotations or repurposed datasets, which restricts scalability and applicability to unconstrained, in-the-wild videos In contrast, \ours provides an automated, multi-operator pipeline and multi-model QA synthesis to deliver a comprehensive benchmark across key human-centric perceptual skills, enabling more precise and scalable evaluation of foundational human video understanding.

\section{More Statistics of \ours}
\label{append:statistics}
\begin{figure}[h!] 
\centering
\includegraphics[width=1.0\linewidth]{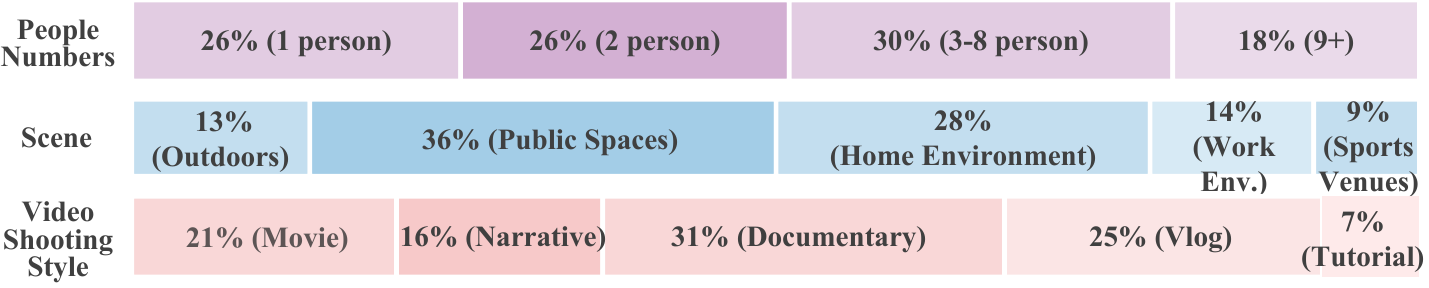}
\caption{The distribution of the number of people, scenes, and video shooting styles in \ours}
\label{fig:humanvbench_details}
\end{figure}

\ours focuses on short video understanding, specifically videos with a duration of 30-second or less. It includes a total of 2475 question instances, with the specific number for each task indicated in \ref{fig:HumanVBench_fig}. The total video duration amounts to 5.2 hours and demonstrates a variety of people, scenes, and video shooting styles, as shown in \ref{fig:humanvbench_details}.

\section{Full Results for 24 Models on 16 Tasks} 
\label{append: Full_Results}

We meticulously select 22 SOTA open-source video MLLMs. 
These included both visual-only models such as 
Qwen-VL series \citep{bai2025qwen2, wang2024qwen2},
InternVL series \citep{chen2024internvl, chen2025expandingperformanceboundariesopensource},
VideoLLaMA3, InternVideo2.5 \citep{wang2025internvideo25empoweringvideomllms}, ViLAMP \citep{cheng2025scaling},LLaVA-Video \citep{zhang2025llavavideo},
ShareGPT4Video, CogVLM2-video \citep{hong2024cogvlm2}, LLaVA-One-Vision \citep{li2024llavaonevisioneasyvisualtask}, Chat-UniVi \citep{jin2024chatuniviunifiedvisualrepresentation}, VILA \citep{lin2024vila}, VideoChat \citep{li2023videochat}, and audio-visual models capable of analyzing both visual and audio inputs, such as the Qwen2.5-Omni \citep{xu2025qwen25omnitechnicalreport}, MIO \citep{wang2025miofoundationmodelmultimodal}, Video-LLaMA series \citep{zhang2023video, cheng2024videollama}
and generalist MLLMs like ImageBind-LLM \citep{girdhar2023imagebind}, ChatBridge \citep{zhao2023chatbridge}, and OneLLM \citep{han2024onellm}.

\begin{table*}[h]
\belowrulesep=0pt
\aboverulesep=0pt
\renewcommand{\arraystretch}{1.0} 
\scriptsize 
\setlength{\tabcolsep}{3.7pt} 
\setlength{\abovecaptionskip}{0.2cm}  
\setlength{\belowcaptionskip}{-0.2cm} 
\caption{
Results on \ours\ with 18 visual-only MLLMs, 7 audio-visual MLLMs, and 5 proprietary models.  ``Random'' denotes random guessing, and ``Human'' indicates human-level performance. Task acronyms are defined in Figure \ref{fig:HumanVBench_fig}. For each task, the best overall results are shown in bold, while the best open-source model results are underlined. ``-'' means the model recognizes its lack of required capabilities for the task and thus refuses to answer. The results for all open-source models are averaged over five runs with different random seeds. } 
\begin{tabular}{l|llll|>{\columncolor{gray!20}}l|llll|>{\columncolor{gray!20}}l|llll|>{\columncolor{gray!20}}l|llll|>{\columncolor{gray!20}}l}
\toprule
&\multicolumn{5}{c|}{\textbf{Emotion Perception}} & \multicolumn{5}{c|}{\textbf{Person Recognition}} & \multicolumn{5}{c|}{\textbf{Behavior Analysis}} & \multicolumn{5}{c}{\textbf{Speech-Visual Alignment}} \\

 \textbf{Models}& \textbf{ER} & \textbf{ETA} & \textbf{AR} & \textbf{EIC} & \textbf{Avg} & \textbf{T2H} & \textbf{H2T} & \textbf{HC} & \textbf{ATD} & \textbf{Avg} & \textbf{BTA} & \textbf{BCA} & \textbf{AST} & \textbf{TSA} & \textbf{Avg} & \textbf{AVSM} & \textbf{ASD} & \textbf{AVAD} & \textbf{SCM} & \textbf{Avg} \\
\toprule
Random & 24.6 & 24.7 & 25.0 & 23.1 & 24.4 & 27.9 & 24.6 & 23.1 & 25.0 & 25.2 & 23.0 & 23.6 & 25.0 & 20.0 & 22.9 & 42.8 & 23.6 & 33.3 & 25.0 & 31.2 \\
\hline
ViLA & 32.7 & 24.8 & 28.7 & 21.4 & 26.9 & 50.4 & 27.0 & 40.4 & 20.2 & 34.5 & 33.0 & 48.6 & 33.0 & 51.4 & 41.5 & 47.4 & 23.9 & 34.3 & 18.6 & 31.1 \\
Video-LLaVA & 18.3 & 18.8 & 26.2 & 8.0 & 17.8 & 27.9 & 40.6 & 28.3 & 31.7 & 32.1 & 30.8 & 32.5 & 27.0 & 40.7 & 32.8 & 50.0 & 28.4 & 34.3 & 21.2 & 33.5 \\
Chat-Univi & 26.4 & 15.0 & 10.9 & 16.5 & 17.2 & 29.1 & 27.0 & 17.8 & 19.8 & 23.4 & 27.9 & 39.4 & 14.0 & 6.7 & 22.0 & 42.8 & 20.6 & 26.3 & 18.6 & 27.1 \\
VideoChat2-IT & 33.5 & 29.1 & 39.6 & 26.8 & 32.3 & 20.1 & 47.3 & 11.2 & 26.7 & 26.3 & 45.8 & 43.7 & 34.0 & 24.9 & 37.1 & 43.4 & 27.7 & 31.4 & 23.0 & 31.4 \\
InternVL2 & 36.1 & 31.3 & 40.2 & 30.4 & 34.5 & 70.4 & 61.2 & 37.2 & 20.8 & 47.4 & 49.8 & 52.3 & 38.0 & 33.8 & 43.5 & 51.7 & 55.0 & 33.6 & 25.5 & 41.5 \\
Qwen-VL2 & 38.0 & 35.1 & 42.7 & 37.5 & 38.3 & 79.3 & 72.7 & 43.4 & 20.8 & 54.1 & 47.8 & 55.6 & 32.0 & 51.4 & 46.7 & 50.7 & 56.1 & 31.4 & 23.7 & 40.5 \\
CogVLM2-Video & 38.4 & 29.9 & 36.0 & 25.0 & 32.3 & 42.5 & 51.5 & 31.1 & 40.0 & 41.3 & 41.9 & 50.3 & 34.0 & 14.1 & 35.1 & 59.2 & 38.7 & 30.7 & 17.9 & 36.6 \\
VideoLLaMA3 & 34.6 & 29.9 & 48.0 & 46.4 & 39.7 & 90.3& 67.9 & 44.9 & \underline{71.0}& \underline{68.5}& 41.5 & 56.3 & \underline{56.4}& \underline{69.0}& 55.8& 64.6 & 63.1& \underline{34.3} & 17.9 & 45.0 \\
LLaVAOneVision & 36.9 & 31.3 & \textbf{\underline{62.8}}& 28.6 & 39.9 & 67.0 & 61.2 & 49.1 & 10.0 & 46.8 & 45.5 & 55.6 & 37.0 & 47.5 & 46.4 & 52.6 & 51.6 & 35.0 & 26.9 & 41.5 \\
InternVideo & 35.7 & \underline{42.9}& 47.6& 35.7 & 40.5 & 64.2& 69.1 & 50.9 & 42.5 & 56.7& 56.1 & 58.3 & 42.0& 55.9 & 53.1 & 59.4& 59.4 & 32.8 & 31.6& 41.5 \\
InternVL2.5 & 37.3 & 38.1& 54.3 & 37.5& 41.8 & 81.0 & 73.3& 40.6 & 35.8 & 57.7 & 53.0 & 57.6& 41.0 & 62.1& 53.4& 65.1 & 61.3& 32.1 & 15.4 & 43.5 \\
InternVL3 & 38.0 & 34.6 & 61.6 & 35.7 & 42.5 & 83.8 & 74.5 & 46.2 & 45.8 & 62.6 & 53.8 & 60.9 & 54.0 & 61.0 & 57.4 & 63.8 & 58.7 & \underline{34.3} & 26.5 & 45.8 \\
InternVL3.5 & 43.7 & 40.6 & 61.0 & 44.6 & \underline{47.5} & 87.2 & 79.4 & 46.2 & 16.7 & 57.4 & 66.4 & 61.6 & 49.0 & 58.2 & \underline{58.8} & 63.8 & 63.2 & 31.4 & 23.1& 45.4 \\
Qwen2.5-VL & \underline{43.0} & 30.6 & 35.4 & \textbf{\underline{50.0}}& 39.8& 88.8& 74.5& 50.9& 30.8 & 61.3 & 51.0 & 58.3 & 34.0 & 50.8 & 48.5 & \underline{71.1}& 61.3& 33.6 & 18.8 & 46.2\\
Qwen3-VL & 40.3 & 37.6 & 53.0 & 42.0 & 43.2 & \underline{94.4} & \underline{77.6} & \underline{51.9} & 46.7 & 67.6 & \underline{60.5} & \underline{64.2} & 39.0 & 53.7 & 54.3 & \underline{71.1} & \underline{65.8} & 34.3 & 22.2 & 48.3 \\
ShareGPT4Video & 35.0 & 39.1& 40.9 & 29.5 & 36.1 & 33.0 & 38.8 & 24.5 & 43.3& 34.9 & 32.8 & 43.7 & 39.0 & 16.9 & 33.1 & 44.1 & 31.0 & \underline{34.3}& 25.0 & 33.6 \\
LLaVA-Video & 35.4 & 32.3 & 59.8& 28.6 & 39.0 & 74.3 & 61.8& 44.3 & 26.7 & 51.8 & 41.9 & 57.0& 47.0 & 58.8& 51.2 & 52.0 & 58.7& 32.8 & 39.3& 45.7\\
ViLAMP & 41.4& 29.3 & 48.8 & 31.2 & 37.7& 59.8 & 62.4 & 45.3 & 20.0 & 46.9 & 55.7 & 49.7 & 52.0 & 36.7 & 48.5 & 53.3 & 51.0 & 32.3 & 31.6 & 42.1 \\
\hline
VideoLLaMA & 30.8 & 21.6 & 11.5 & 20.5 & 21.1 & 28.4 & 23.0 & 23.9 & 20.0 & 23.8 & 22.9 & 41.1 & 23.2 & 15.4 & 25.7 & 40.1 & 26.6 & 33.1 & 26.2 & 31.5 \\
VideoLLaMA2.1 & 38.0& 30.6& 47.0& 30.4& 36.5& 34.6 & 52.7& 41.5& 16.7 & 36.4 & 53.8& 60.3& 27.0 & 33.3 & 43.6 & 44.0 & 31.6 & 32.1 & 23.7 & 32.9 \\
MIO & 29.3 & 20.3 & 29.3 & 10.7 & 22.4 & 4.5& 18.2 & 10.4 & 23.3 & 14.1& 22.9 & 39.7 & 29.0 & 19.2 & 27.7 & 42.1 & 31.6 & 16.1 & 19.6 & 27.4 \\
ImageBind-LLM & 21.3 & 22.4 & 25.0 & 11.6 & 20.1 & 23.5 & 13.3 & 23.8 & 19.5 & 20.0 & 15.4 & 32.4 & 24.0 & 23.3 & 23.8 & 45.0 & 25.1 & 28.9 & 22.9 & 30.5 \\
ChatBridge & 28.9 & 18.0 & 30.2 & 16.1 & 23.3 & 31.4 & 30.9 & 23.7 & 8.7 & 23.7 & 24.1 & 43.7 & 41.0 & 16.9 & 31.4 & 43.7 & 25.3 & 30.4 & 24.4 & 31.0 \\
OneLLM & 27.0 & 26.1 & 34.1 & 7.1 & 23.6 & 29.0 & 36.4 & 20.6 & 27.8 & 28.5 & 24.9 & 49.0 & 23.0 & 22.4 & 29.8 & 43.4 & 26.4 & 29.5 & 23.2 & 30.6 \\
Qwen2.5-Omni & 42.2 & 30.1 & 36.6 & 33.0 & 35.5 & 56.4 & 64.2& 41.5 & 15.8 & 44.5 & 45.8 & 55.0& 28.0 & 24.3 & 38.3 & \underline{71.1}& 48.4& 27.0& \underline{71.8}& \underline{54.6}\\
\hline
GPT-4o (20241120) & 36.5 & 43.6 & 25.6 & 28.5 & 33.6 & 49.2 & 77.0& 47.2 & 32.5 & 50.9& 62.1& 62.9& 52.0 & 71.2& 62.1& - & - & - & - & - \\
GPT-5 (20250807) & 43.7 & 48.9 & 59.8 & 34.8 & 46.8 & 69.8 & 84.8 & 64.2 & 58.3 & 69.5 & 73.1 & 66.9 & 64.0 & 65.0 & 67.3 & - & - & - & - & - \\
Gemini-1.5-Pro & 49.4& 51.9& 53.0& 49.1& 50.9& 87.1& 73.9& 52.8& 71.7& \textbf{71.4} & 53.4& 60.3& 54.0& \textbf{75.1} & 60.7& 90.1& 76.8& 66.4& 84.6& 79.5\\
Gemini-2.5-flash & 52.1 & \textbf{57.1}& 59.1 & 43.8 & 53.0 & 92.2 & 77.6 & 61.3 & 71.7 & 75.7 & 65.6 & 68.9 & 58.0 & 67.2 & 64.9 & \textbf{98.7}& \textbf{80.6}& 57.7 & \textbf{99.1}& 84.0 \\
Gemini-2.5-Pro & \textbf{54.4}& 54.1 & 53.0 & \textbf{50.0}& \textbf{52.9}& \textbf{95.0}& \textbf{84.8}& \textbf{69.8}& \textbf{84.2}& \textbf{83.5}& \textbf{79.0}& \textbf{72.2}& \textbf{63.0}& 68.4 & \textbf{70.7}& 96.7 & 78.7 & \textbf{72.3}& 98.3 & \textbf{86.5}\\
\hline
Human & 87.6 & 85.0 & 87.8 & 78.0 & 84.6 & 98.9 & 84.1 & 92.5 & 78.3 & 88.5 & 86.7 & 84.5 & 88.6 & 88.1 & 87.0 & 96.0 & 96.1 & 87.0 & 98.3 & 94.4 \\
\bottomrule
\end{tabular}
\label{table:overall_appendix}
\end{table*}

\begin{figure}[h!] 
\centering
\includegraphics[width=1.0\linewidth]{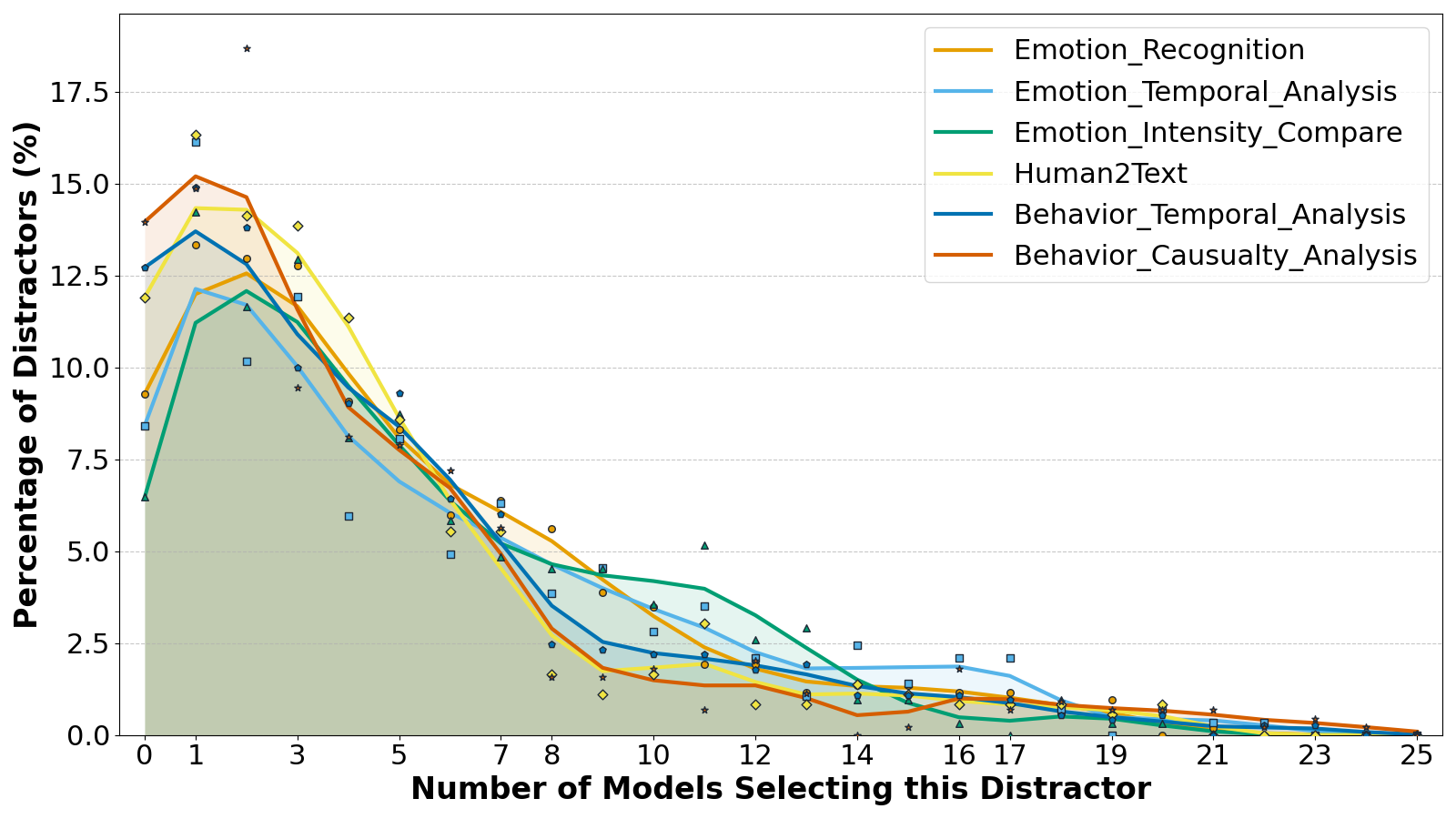}
\caption{Histogram of smoothed distractor selection frequency by 25 open-source MLLMs across six tasks, generated via Distractor-Included Synthesis QA pipeline.}
\label{fig:distractor_analysis}
\end{figure}

\section{Distractor Quality Analysis}
\label{append:distractor_evaluation}
To quantitatively assess the quality of the distractors generated by our synthesis pipeline, we compute the selection frequency of every distractor across the 25 open-source MLLMs evaluated in our study, and visualize the resulting histogram distribution (Figure \ref{append:distractor_evaluation}). The analysis reveals two distinctive properties. First, the distractors exhibit exceptionally high effectiveness: the number of items at X=0 is extremely small, and on six tasks, approximately 90\% of distractors are selected by at least one model. This demonstrates that our “model-induced error” generation strategy successfully avoids trivial or obviously incorrect options, compelling models to perform fine-grained reasoning over all choices. Second, the distribution forms an ideal discriminatory pattern, with the majority of distractors selected by 1 to 8 models and a clear long-tail structure. This indicates that most distractors effectively attract and differentiate about 4\%–32\% of the models, reflecting high-quality deceptiveness and strong discriminatory power. The long-tail pattern further suggests a healthy difficulty gradient: the benchmark neither becomes too easy (which would render distractors ineffective) nor too difficult (which would cause random guessing).

Overall, the combination of a low-frequency peak and a long-tailed distribution provides strong evidence that our distractors are both well-formed and discriminative, forming a solid foundation for a reliable evaluation benchmark.

Moreover, model accuracy results strongly validate this effect: performance follows the expected capability hierarchy—commercial frontier models (e.g., Gemini-2.5-Pro, Gemini-1.5-Pro) outperform advanced open-source 7B models (e.g., Qwen3-VL, InternVL3.5), which in turn surpass earlier models of similar scale (e.g., Video-LLaVA, ShareGPT4Video). This gradient confirms that the distractors successfully penalize weaker models while stronger models remain more resilient, further showcasing the benchmark’s discriminatory strength and reliability.

\section{Modality Ablation in VideoLLaMA2} 
\label{append:ModalityAblation}
\begin{table}[h]
\centering 
\belowrulesep=0pt
\aboverulesep=0pt
\renewcommand{\arraystretch}{1.0} 
\footnotesize 
\setlength{\tabcolsep}{3pt} 
\setlength{\abovecaptionskip}{0.2cm}  
\setlength{\belowcaptionskip}{-0.2cm} 
\caption{The performance of VideoLLaMA2's vision-only version (VideoLLaMA2-7B-16F), and the audio-visual version (VideoLLaMA2.1-AV), on \ours, based on different modal inputs (A for Audio, V for Visual).} 
\begin{tabular}{l|l|llll|>{\columncolor{gray!20}}l}
\toprule

\multirow{2}{*}{\textbf{Models}} & \multirow{2}{*}{\makecell{\textbf{Input} \\ \textbf{Modal}}} &  \multicolumn{5}{c}{\textbf{Speech-Visual Alignment}} \\

 & & \textbf{AVSM} & \textbf{ASD} & \textbf{AVAD} & \textbf{SCM} & \textbf{Avg} \\
\toprule
\multicolumn{2}{c|}{Random} & 42.8 & 23.6 & 33.3 & 25.0 & 31.2
\\
\hline
Video-LLaMA2.1-AV & A, V & 44.0 & 31.6 & 32.1 & 23.7 & 32.9
\\
Video-LLaMA2.1-AV & V & 43.4 & 29.0 & 32.8 & 18.8 & 31.0
\\
VideoLLaMA2-7B-16F & V & 47.4 & 38.1 & 32.1 & 15.4 & 33.3\\
\bottomrule
\end{tabular}
\label{table:videollama2_series}
\end{table}

Despite audio-visual MLLMs processing audio data, they perform at random-guess levels on AVSM and ASD tasks, underperforming relative to many vision-only models that rely solely on lip movement analysis. This raises the question: does the poor performance stem from limitations in visual analysis (e.g., lacking lip-reading ability) or from the interference of audio input? To explore this, we conducted ablation experiments using the VideoLLaMA2 model series, chosen for its open-source availability of both vision-only and audio-visual variants.

As shown in the Table \ref{table:videollama2_series}, VideoLLaMA2-7B-16F (vision-only) exhibits only a slight advantage over Video-LLaMA2.1-AV (audio-visual) on AVSM and ASD tasks, yet still lags far behind vision-only models such as VideoLLaMA3 and InternVL2.5 (Table \ref{table:overall_appendix}). This indicates that VideoLLaMA2-7B has inherently poor lip-reading capability, which further implies that the audio-visual variant (Video-LLaMA2.1-AV) also suffers from limited visual lip-reading ability. Such limitations constrain its upper-bound performance in speech-visual alignment tasks. On the other hand, Video-LLaMA2.1-AV shows no significant performance advantage when utilizing audio information compared to its vision-only counterpart. This suggests that vocal information is not effectively leveraged, likely due to insufficient speech parsing capability in video MLLMs and inadequate understanding of cross-modal associations between audio and visual content.

\section{Impact of Frame Sampling}
\label{append:rebuttal_frame}
\begin{table}[h]
\centering
\footnotesize
\makeatletter
\setlength{\aboverulesep}{0pt}
\setlength{\belowrulesep}{0pt}
\makeatother
\caption{Impact of different frame sampling strategies on HUMANVBENCH. We report the accuracy (\%) and total inference time (minutes) for InternVL3.5 and LLaVA-Video. Denser sampling beyond 8 frames leads to diminishing returns and increased computational overhead.}
\label{tab-frame_sampling}
\begin{tabular}{l|ccc|ccc} 
\toprule
\textbf{Model} & \multicolumn{3}{c|}{\textbf{InternVL3.5}} & \multicolumn{3}{c}{\textbf{LLaVA-Video}} \\
\midrule
Sampling Rate & 4f & 8f & 16f & 4f & 8f & 16f \\ \midrule
Emotion (4 tasks)      & 45.2 & 46.1 & \textbf{47.2} & 36.9 & \textbf{39.4} & 38.8 \\
Person Recog. (4 tasks)& 55.0 & 57.6 & 56.5 & 52.2 & 53.2 & 53.9 \\
Behavior (4 tasks)     & 57.0 & 58.6 & 59.9 & 51.3 & 52.0 & 53.0 \\
AV-Sync (4 tasks)      & 44.8 & 46.0 & 45.8 & 42.3 & 44.0 & 43.5 \\ \midrule
\textbf{Overall Avg.}  & 50.5 & 52.1 & 52.3 & 45.7 & 47.2 & 47.3 \\ \midrule
\textbf{Total Time} (min.)      & 28.3 & 39.2 & 62.4 & 28.7 & 40.9 & 62.3 \\ \bottomrule
\end{tabular}%
\end{table}
To investigate the effect of temporal resolution on HUMANVBENCH, we evaluate representative MLLMs (InternVL3.5 and LLaVA-Video) across 4, 8, and 16 frames. As shown in Table \ref{tab-frame_sampling}, while denser sampling generally yields performance gains, we observe significant diminishing returns. Specifically, for InternVL3.5, increasing from 8 to 16 frames results in only a marginal $0.26\%$ overall improvement while incurring a $1.6\times$ increase in latency (from 39.2 to 62.4 minutes). Notably, in tasks like Person Recognition, 8-frame sampling even outperforms 16-frame sampling, suggesting that excessive frames may introduce redundant temporal noise. These findings underscore the importance of efficient temporal modeling over brute-force dense sampling.

\section{Impact of Human Density on Model Performance}
\label{append:rebuttal_human_density}
\begin{table}[h]
\centering
\belowrulesep=0pt
\aboverulesep=0pt
\renewcommand{\arraystretch}{1.1} 
\footnotesize
\setlength{\tabcolsep}{6pt} 
\setlength{\abovecaptionskip}{0.2cm}
\setlength{\belowcaptionskip}{-0.2cm}
\caption{Performance across varying human densities. Human density is categorized as \textbf{Low} (1--3 persons), \textbf{Med} (4--8), and \textbf{High} ($>8$). Accuracy (\%) is reported for Strong and Mid-tier groups.}
\label{tab:density_analysis}
\begin{tabular}{l|ccc|ccc} 
\toprule
\textbf{Task Category} & \multicolumn{3}{c|}{\textbf{Strong Models}} & \multicolumn{3}{c}{\textbf{Mid-tier Models}} \\
\cmidrule(r){2-4} \cmidrule(l){5-7}
(Human Density) & Low & Med & High & Low & Med & High \\ 
\toprule
\rowcolor{gray!20} Overall Average & 63.5 & 57.2 & 47.2 & 37.3 & 36.8 & 35.6 \\
\midrule
Emotion Perception & 49.0 & 50.8 & 50.0 & 35.5 & 34.9 & 39.5 \\
Human Recognition  & 78.1 & 65.1 & 36.1 & 42.4 & 38.0 & 31.9 \\
Behavior Analysis  & 61.6 & 56.0 & 55.6 & 33.8 & 40.2 & 35.3 \\
Speech Alignment   & 65.1 & 57.0 & ---  & 37.3 & 34.2 & ---  \\
\bottomrule
\end{tabular}
\end{table}

We evaluate two model tiers—Strong (InternVL3.5, Qwen2-VL-72B, VideoLLaMA3) and Mid-tier (InternVL2, CogVLM2-Video, Qwen2-VL-7B)—to benchmark robustness against varying human densities. As shown in Table \ref{tab:density_analysis}, a capability collapse occurs in autonomous localization (e.g., Human Recognition), where Strong models' accuracy plunges from 78.1\% to 36.1\% in dense scenes. Conversely, performance in visually-grounded tasks (e.g., Emotion Perception, where targets are pre-marked by bounding boxes) remains stable. This indicates that multi-entity interference primarily hinders autonomous spatial-temporal grounding—the ability to isolate a specific individual from a crowd—rather than the high-level reasoning itself.

While Strong models maintain a performance lead in all scenarios, this margin narrows in dense environments; for instance, the Overall Average gap between the two tiers shrinks from 26.2\% to 11.6\%. This underscores the need for density-invariant grounding to overcome interference in crowded scenes. Future work should prioritize higher-resolution attention mechanisms to maintain precise focus on specific targets amidst increasing environmental complexity.

\section{Model Evaluation Implementation}
\label{append:eval_detail}
\textbf{Prompt.}
In order to facilitate the statistical model to answer the results, following common practices used in MLLM evaluations, we adopt the following prompt to guide the MLLM to output option letters: ``
\textcolor{gray}{\textit{Select the best answer to the following multiple-choice question based on the video. Respond with only the letter of the correct option. \textless{Question-choices}\textgreater \quad Only answer best answer's option letter. Your option is: }
}
''.
\textbf{Evaluation Environments.}
All evaluation experiments for open-source models were conducted on a single NVIDIA L20 GPU with an inference batch size of 1.

\textbf{Baseline Configurations and Runtime Statistics.}
Table \ref{table:model_config} shows the scale, parameter settings, and costs (including memory usage and end-to-end testing time) for each model on \ours. 
All hyperparameter settings follow the default configurations of these open-source works.

\begin{table}[h!]
\centering
\scriptsize
\setlength{\tabcolsep}{2pt}
\renewcommand{\arraystretch}{1.2}
\begin{tabular}{lcccccc}
\toprule
\textbf{Model} & \textbf{Time (min)} & \textbf{top\_p} & \textbf{top\_k} & \textbf{num\_beams} & \textbf{temp.} & \textbf{frames}\\ 
\midrule
Chat-UniVi (7B)       & 40& 1    & 50 & 1 & 0.2 & 1 f/s \\
CogVLM2-Video (8B)    & 37& 0.1  & 1  & 1 & 0.2 & 1 f/s \\
Video-LLaVA (7B)    & 49& 1    & 50 & 1 & 1   & 8 f \\

LLaVA-OneVision (7B) & 49& 1    & 50 & 1 & 1   & 8 f \\
PLLaVA (7B)        & 41& 0.9  & 50 & 1 & 1   & 16 f \\
ShareGPT4Video (8B)   & 50& 0.9  & 50 & 1 & 1   & 16 f \\
Otter-V (7B)         & 62& 1    & 50 & 3 & 1   & 16 f \\
VideoChat2-IT (7B)    & 53& 0.9  & 50 & 1 & 1   & 16 f \\
 InternVL2 (7B)& 44& 1& 52& 1& 1.0&8 f\\
 InternVL2.5 (7B)& 31& 1& 52& 1& 1.0&8 f\\
 InternVL3 (7B)& 186 & 1& 52& 1& 1.0& 64 f\\
 InternVL3.5 (7B)& 107& 1& 52& 1& 1.0& 32 f\\
 Qwen2-VL (7B)& 45& 0.001& 1& 1& 0.1&2 f/s\\
 Qwen2.5-VL (7B)& 35& 0.001& 1& 1& 0.1&2 f/s\\
 Qwen3-VL (7B)& 34 & 0.001& 1& 1& 0.1&2 f/s\\
 LLaVA-Video (7B)& 182& 0.8& 20& 1& 0.7&64 f\\
 Video-LLaMA3 (7B)& 90& 0.8& 20& 1& 0.7&1 f/s\\
 InternVideo2.5 (7B) & 170 & 1 & 50 & 1 & 1 & 128f \\
 ViLAMP (7B) & 29 & 0.8 & 20 & 1 & 0.7 & 1 f/s\\
\midrule
Video-LLaMA (7B)       & 40& 1    & 50 & 2 & 1   & 8 f \\
Video-LLaMA2.1 (7B)    & 31& 0.9  & 50 & 1 & 0.2 & 8 f\\
ImageBind-LLM (7B)    & 77& 1    & 50 & 1 & 1   & 15 f \\
ChatBridge (13B)       & 29& 1    & 50 & 1 & 0.2 & 4 f \\
OneLLM (7B)           & 130& 0.75 & 50 & 1 & 0.1 & 15 f \\
MIO (7B) & 88 & 0.7 & 0 & 1 & 1 & FPS/10 \\
Qwen2.5\_Omni (7B) & 91 & 1 & 50 & 1 & 1 & 2 f/s \\
\midrule
GPT-4o (20241120)  & 191& 1    & -  & - & 1   & FPS/10   \\
Gemini-1.5-Pro  & 254& 1    & 40  & - & 0.9   & API   \\
\bottomrule
\end{tabular}
\caption{Model configuration and runtime statistics evaluated on \ours (one pass for all provided test samples).}
\label{table:model_config}
\end{table}

\section{Definitions and Examples for Each Task}
\label{append:task_example}
\subsection{Emotion Perception}
\begin{figure}[h!]
\vspace{-1.0em}
    \centering
    \includegraphics[width=1\columnwidth]{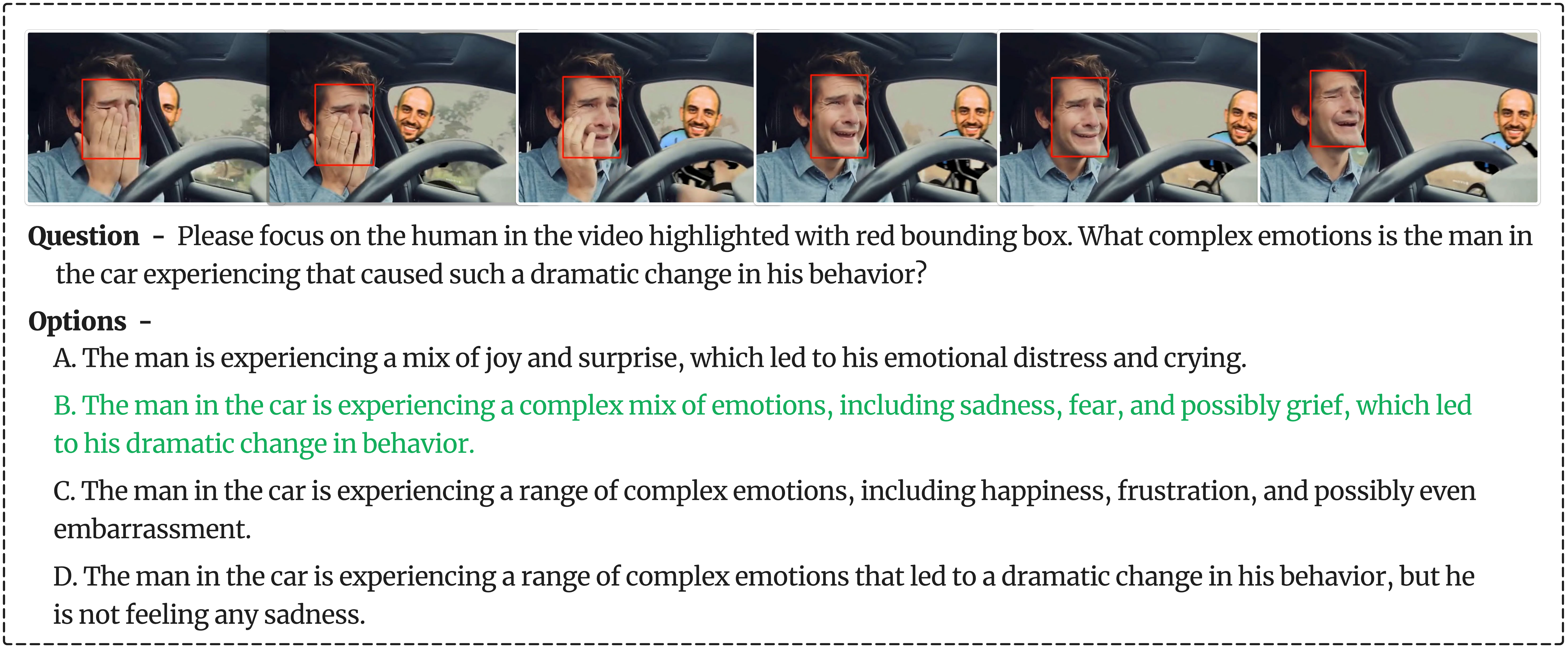}
    \vspace{-0.5em}
    \caption{Example of Emotion Recognition task.}
    \label{fig:Example_of_Emotion_Recognition}
    \vspace{-1em}
\end{figure}

\textbf{Emotion Recognition} aims to judge the overall emotional state of the person highlighted by a red bounding box in the video. An example is shown in Figure \ref{fig:Example_of_Emotion_Recognition}.

\begin{figure}[h!]
\vspace{-1.0em}
    \centering
    \includegraphics[width=1\columnwidth]{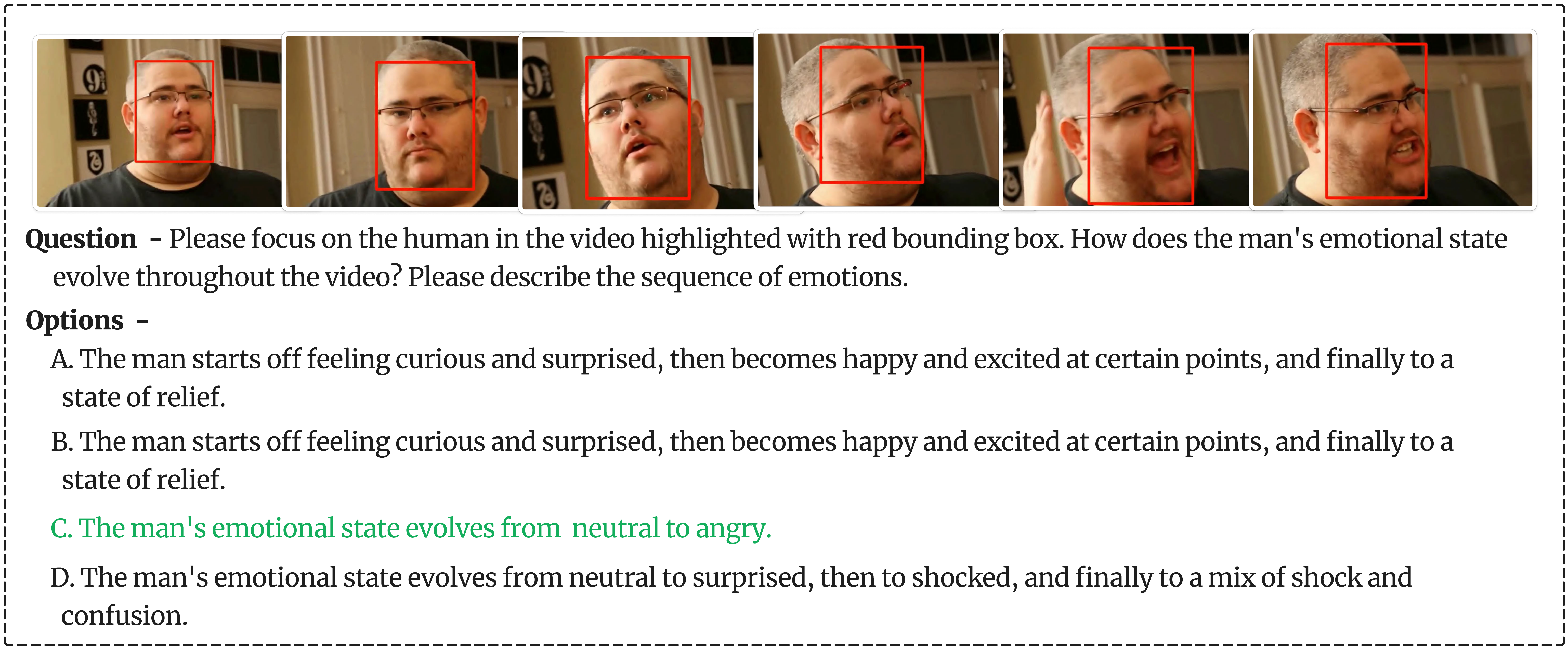}
    \vspace{-0.5em}
    \caption{Example of Emotion Temporal Analysis task.}
    \label{fig:Example_of_Emotion_Temporal_Analysis}
    \vspace{-1em}
\end{figure}

\noindent \textbf{Emotion Temporal Analysis} involves analyzing the changes in the emotions of the people highlighted with the red bounding box over time, identifying gradual intensification, diminishment, emotions shifts to test the model's ability to track emotional dynamics. An example is shown in Figure \ref{fig:Example_of_Emotion_Temporal_Analysis}.

\begin{figure}[h!]
    \centering
    \includegraphics[width=1\columnwidth]{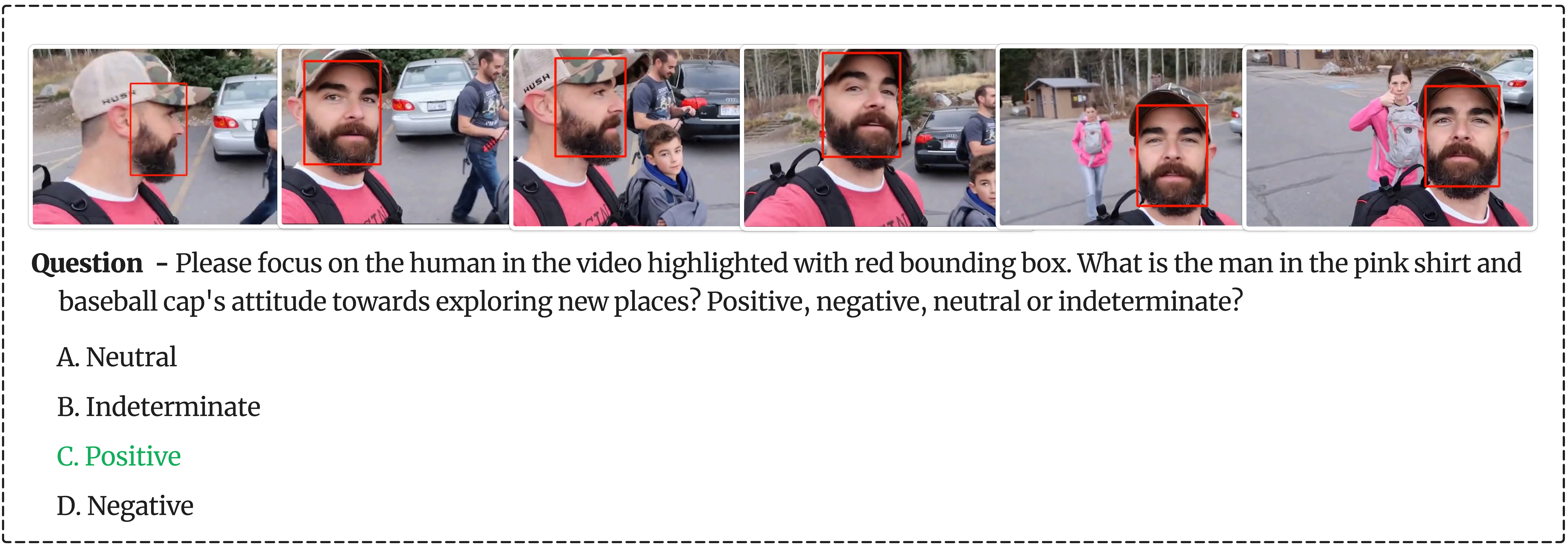}
    \vspace{-0.5em}
    \caption{Example of Attitude Recognition task.}
    \label{fig:Example_of_Attitude_Recognition}
    \vspace{-1em}
\end{figure}

\noindent \textbf{Attitude Recognition} involves inferring a character's attitude towards things, categorized into four fixed options: positive, neutral, negative, and indeterminate. An example is shown in Figure \ref{fig:Example_of_Attitude_Recognition}.

\begin{figure}[h!]
    \centering
    \includegraphics[width=1\columnwidth]{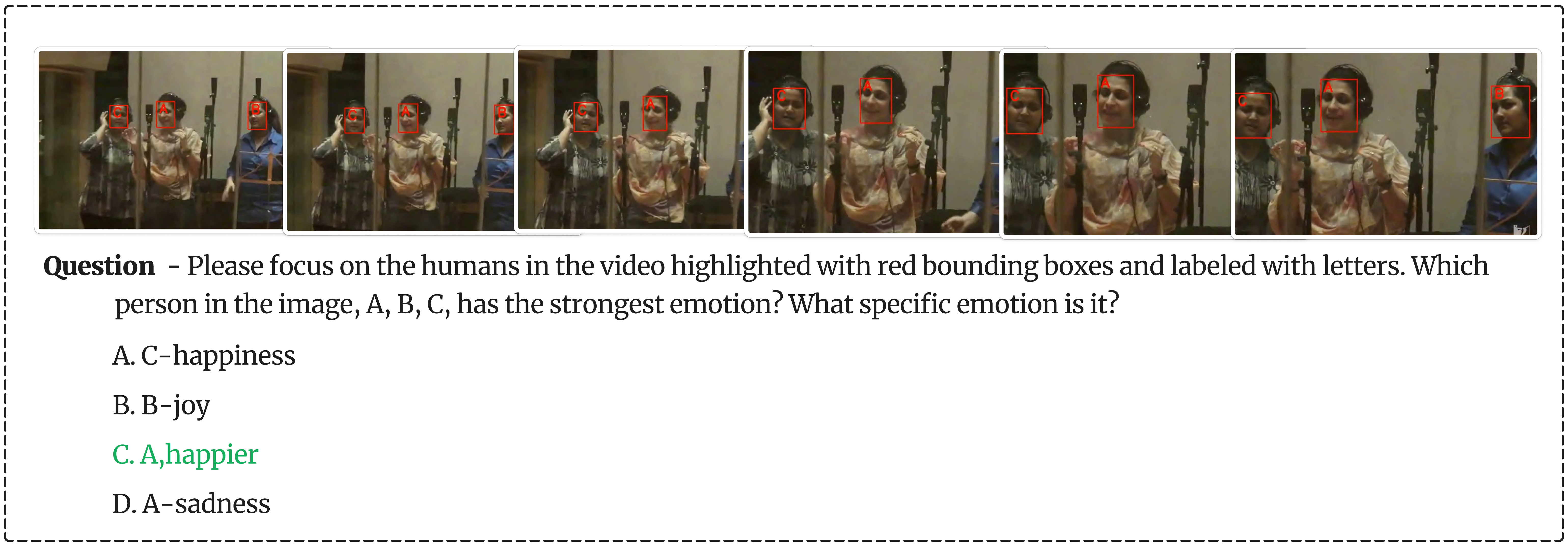}
    \vspace{-0.5em}
    \caption{Example of Emotion Intensity Comparision task.}
    \label{fig:Example_of_Emotion_Intensity_Compare}
    \vspace{-1em}
\end{figure}

\noindent \textbf{Emotion Intensity Comparison} requires compares the emotional intensity differences among various individuals in the video to find the most emotional person, assess whether the model can quantify and differentiate emotional intensity. An example is shown in Figure \ref{fig:Example_of_Emotion_Intensity_Compare}.

\subsection{Person Recognition}

\begin{figure}[h!]
    \centering
    \includegraphics[width=1\columnwidth]{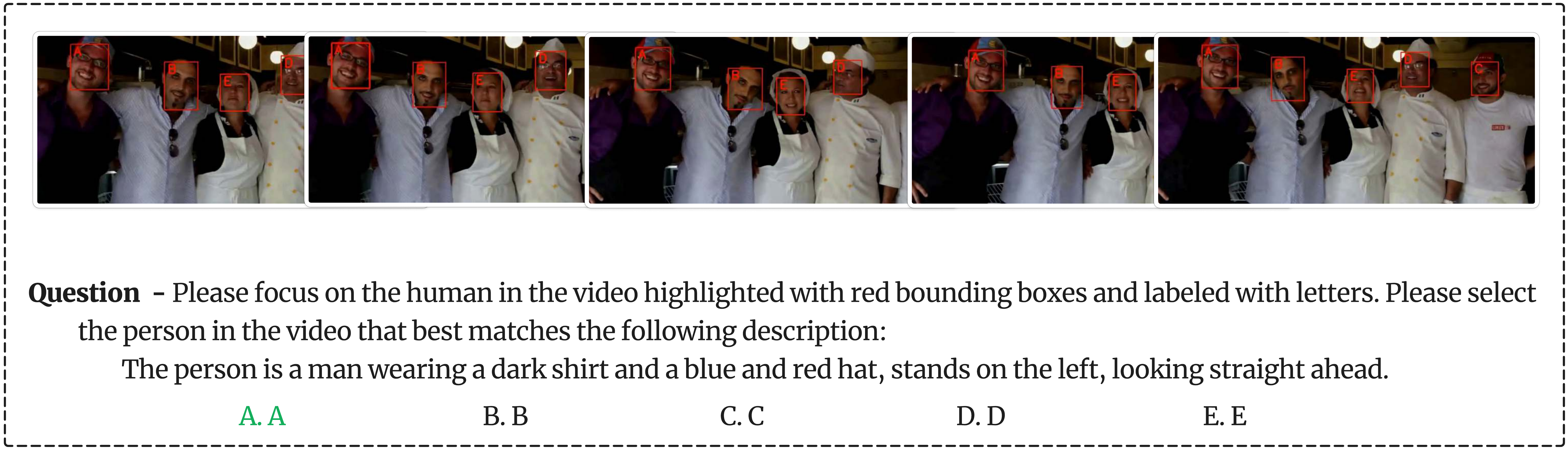}
    \vspace{-0.5em}
    \caption{Example of Text-to-Human task.}
    \label{fig:Example_of_Text2Human}
    \vspace{-1em}
\end{figure}

\noindent \textbf{Text-to-Human} requires the model to identify the specific person in a multi-person video  based on a given text description, to test the model's ability to locate and identify the described person.  An example is shown in Figure \ref{fig:Example_of_Text2Human}.

\begin{figure}[h!]
    \centering
    \includegraphics[width=1\columnwidth]{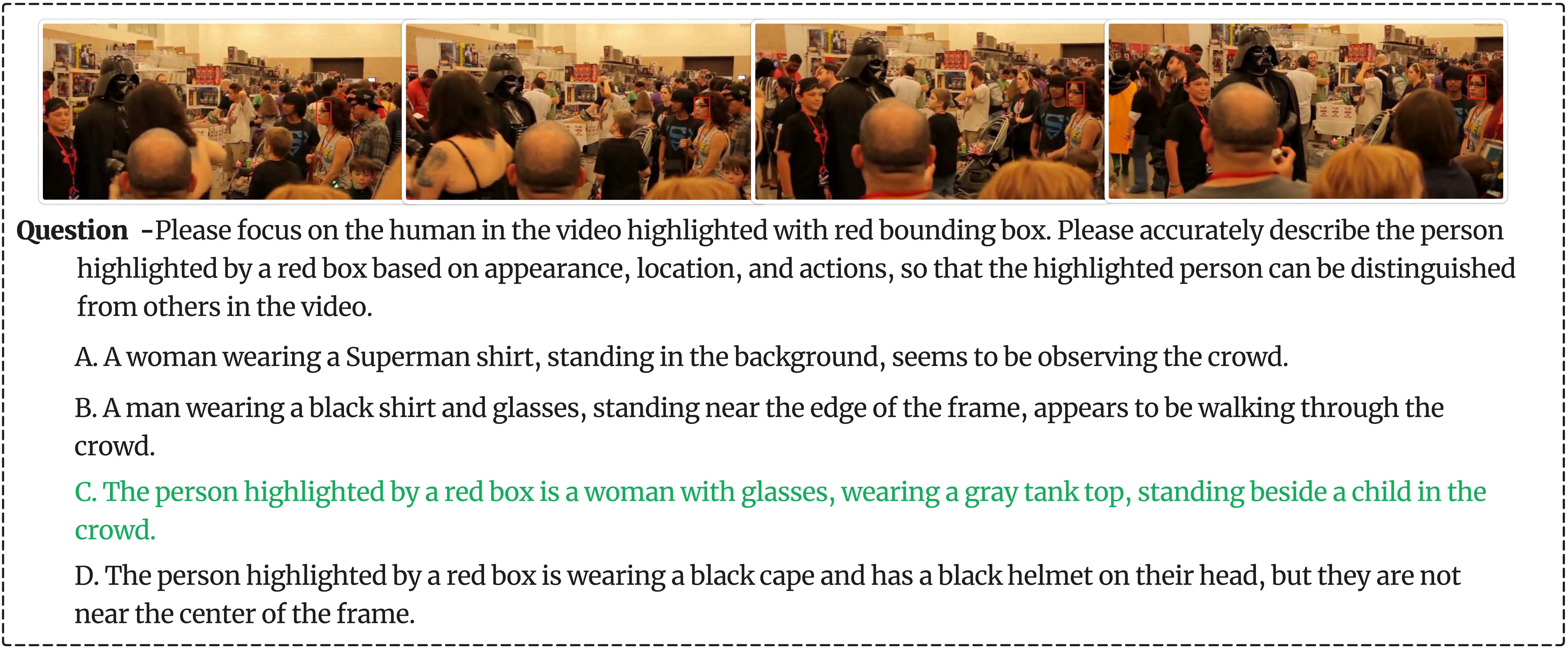}
    \vspace{-0.5em}
    \caption{Example of Human-to-Text task.}
    \label{fig:Example_of_Human2Text}
    \vspace{-1em}
\end{figure}

\noindent \textbf{Human-to-Text} asks the model to choose the most accurate description of the target person in a multi-person video, to ensure that the person is clearly distinguished from others and uniquely identified. This task requires the model to analyze and compare individuals in the video, identifying distinguishing features of the target person, such as appearance, clothing, actions, location, and other characteristics. An example is shown in Figure \ref{fig:Example_of_Human2Text}.\\

\begin{figure}[h!]
    \centering
    \includegraphics[width=1\columnwidth]{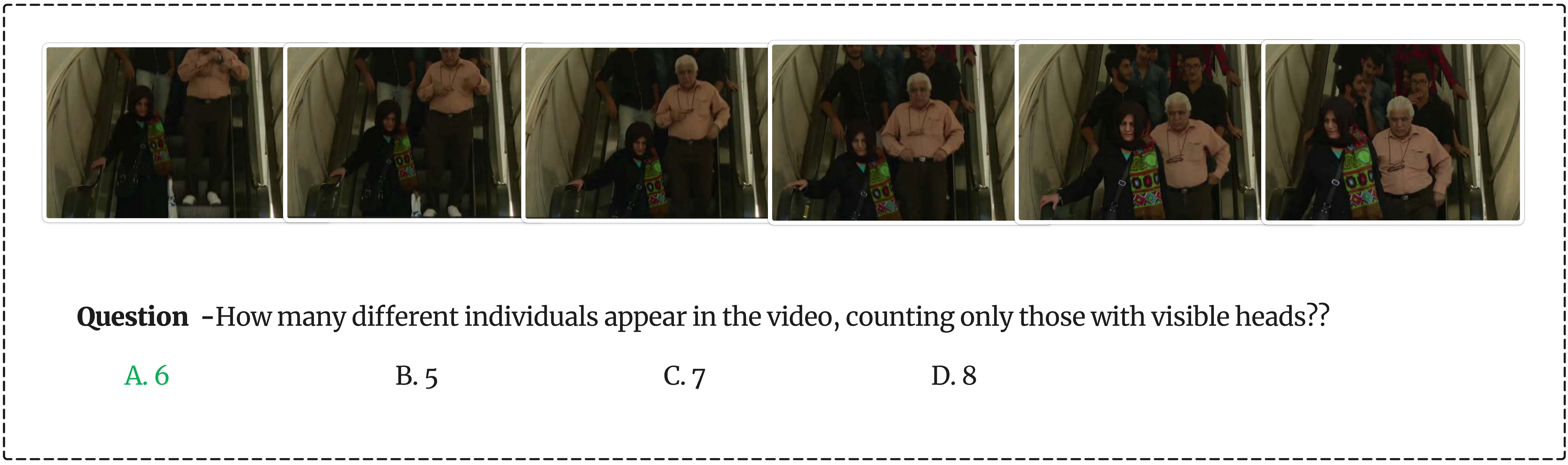}
    \vspace{-0.5em}
    \caption{Example of Human Counting task.}
    \label{fig:Example_of_Human_Count}
    \vspace{-1em}
\end{figure}

\noindent \textbf{Human Counting} requires the model to determine the total number of distinct individuals in the video, testing its capability to detect, track, and accurately count individuals in complex scenes. An example is shown in Figure \ref{fig:Example_of_Human_Count}.\\

\begin{figure}[h!]
    \centering
    \includegraphics[width=1\columnwidth]{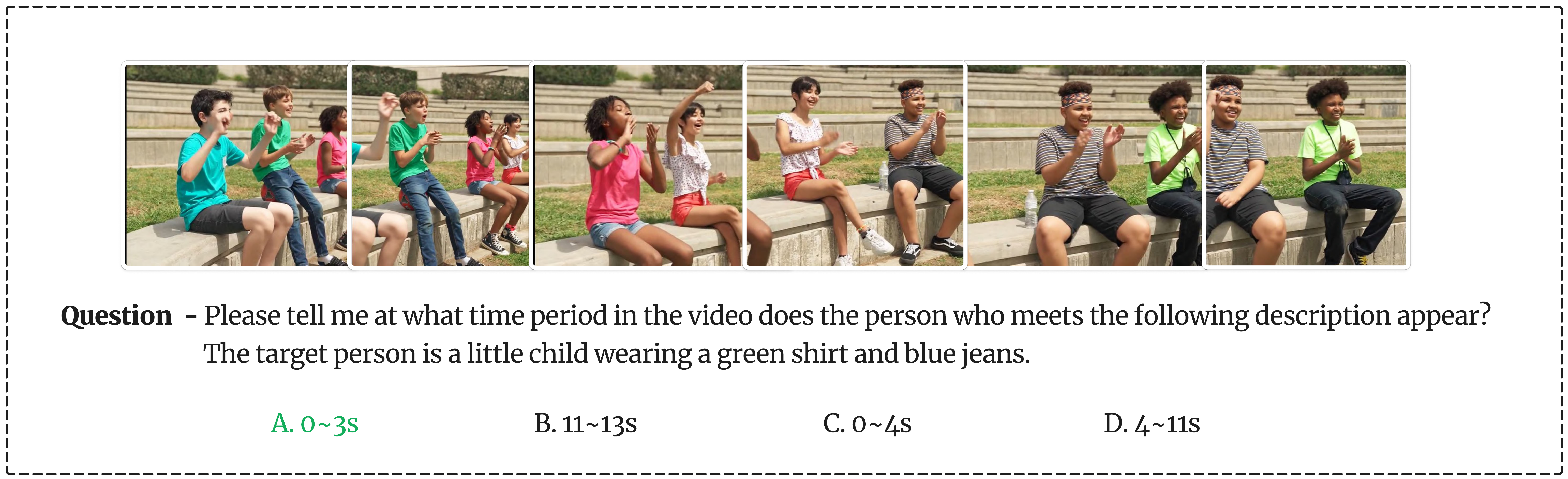}
    \vspace{-0.5em}
    \caption{Example of Appearance Time Detection task.}
    \label{fig:Example_of_Appearance_Time_Detection}
    \vspace{-1em}
\end{figure}

\noindent \textbf{Appearance Time Detection} requires the model to identify the exact time frames when a specified person appears, demanding the ability to precisely mark the start time, end time, and duration of the individual's presence in the video. An example is shown in Figure \ref{fig:Example_of_Appearance_Time_Detection}.

\subsection{Human Behavior Analysis}
\begin{figure}[h!]
    \centering
    \includegraphics[width=1\columnwidth]{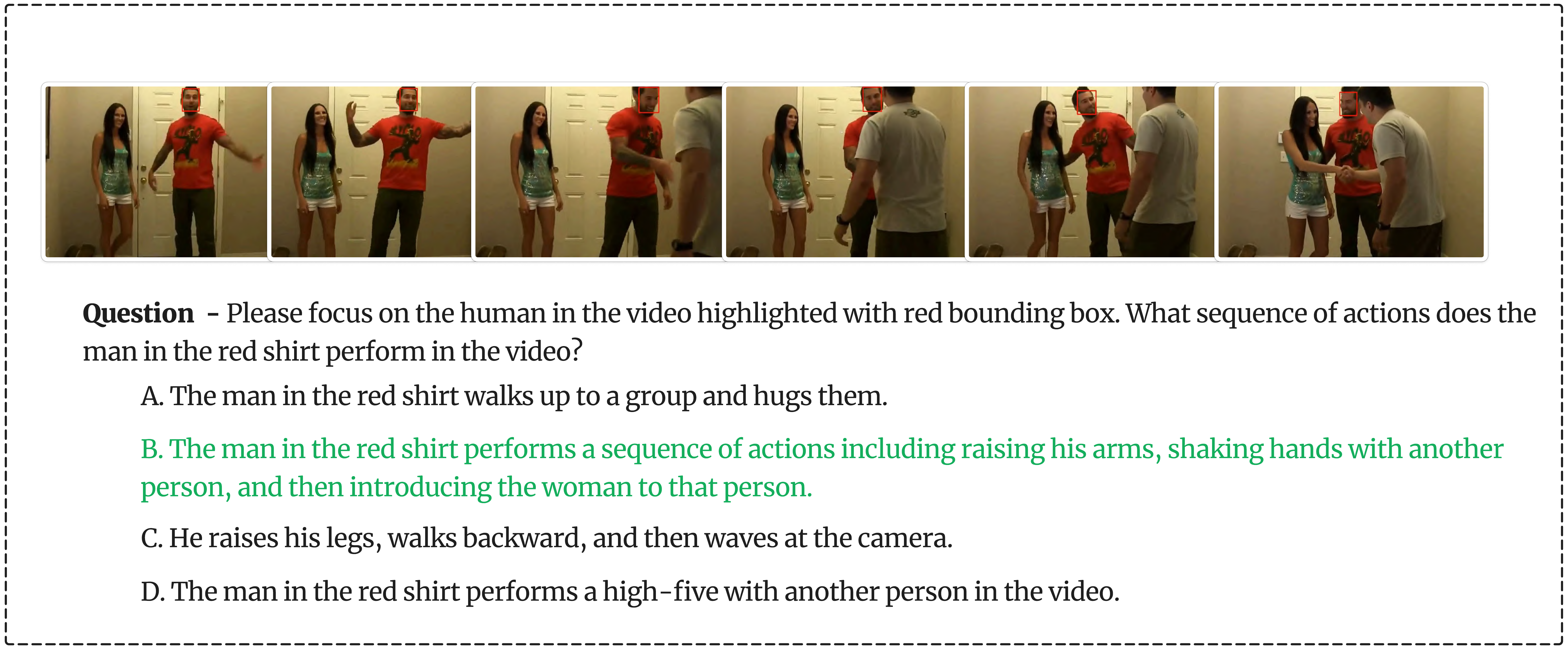}
    \vspace{-0.5em}
    \caption{Example of Behavoir Temporal Analysis task.}
    \label{fig:Example_of_Behavoir_Temporal_Analysis}
    \vspace{-1em}
\end{figure}

\noindent \textbf{Behavior Temporal Analysis} involves analyzing the dynamic changes in a specified person's behavior over time, testing the model's ability to accurately capture and track the temporal characteristics of these changes. An example is shown in Figure \ref{fig:Example_of_Behavoir_Temporal_Analysis}.\\

\begin{figure}[h!]
    \centering
    \includegraphics[width=1\columnwidth]{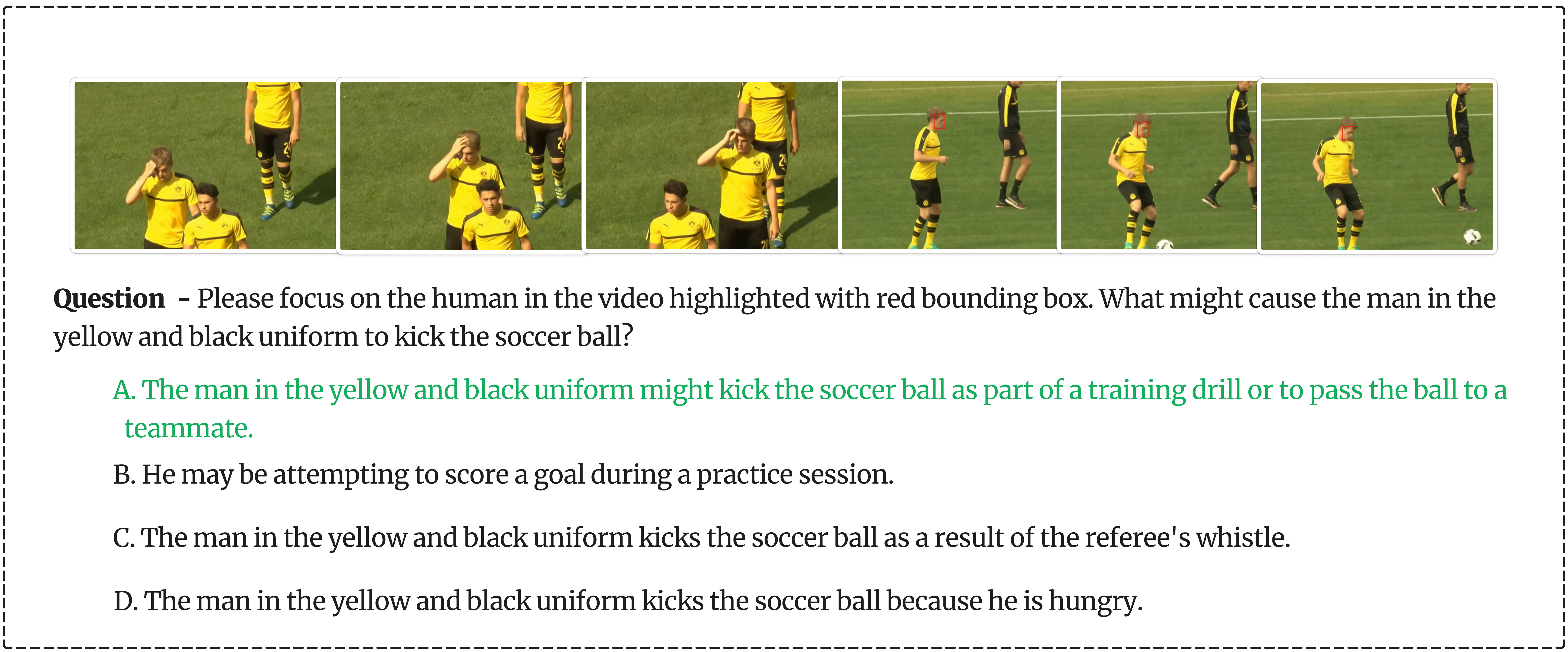}
    \vspace{-0.5em}
    \caption{Example of Behavior Causualty Analysis task.}
    \label{fig:Example_of_Behavior_Causualty_Analysis}
    \vspace{-1em}
\end{figure}

\noindent \textbf{Behavior Causality Analysis} aims to investigate the causal relationships underlying a specific behavior, requiring the model to determine whether a person's behavior in the video is triggered by a particular event or leads to subsequent actions. An example is shown in Figure \ref{fig:Example_of_Behavior_Causualty_Analysis}.\\

\begin{figure}[h!]
    \centering
    \includegraphics[width=1\columnwidth]{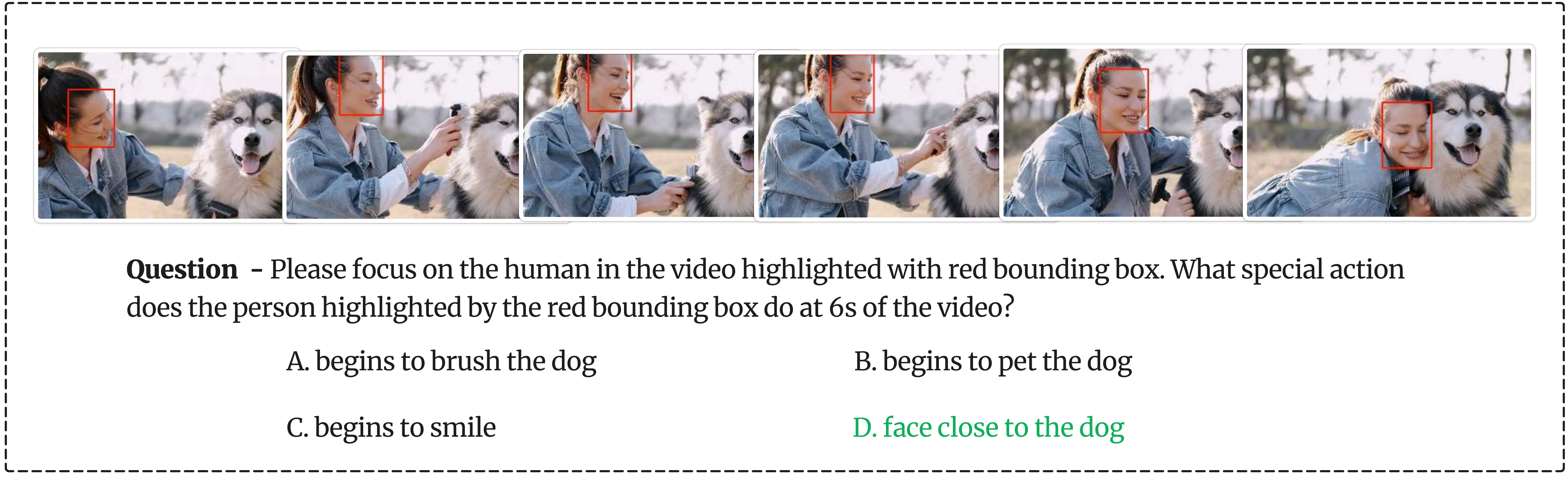}
    \vspace{-0.5em}
    \caption{Example of Action at Specific Time task.}
    \label{fig:Example_of_Action_at_Specific_Time}
    \vspace{-1em}
\end{figure}

\noindent \textbf{Action at Specified Time} asks the model to identify a person's behavior or state at a specific time, testing its ability to accurately determine the person's action or state at the given moment. An example is shown in Figure \ref{fig:Example_of_Action_at_Specific_Time}.\\

\begin{figure}[h!]
    \centering
    \includegraphics[width=1\columnwidth]{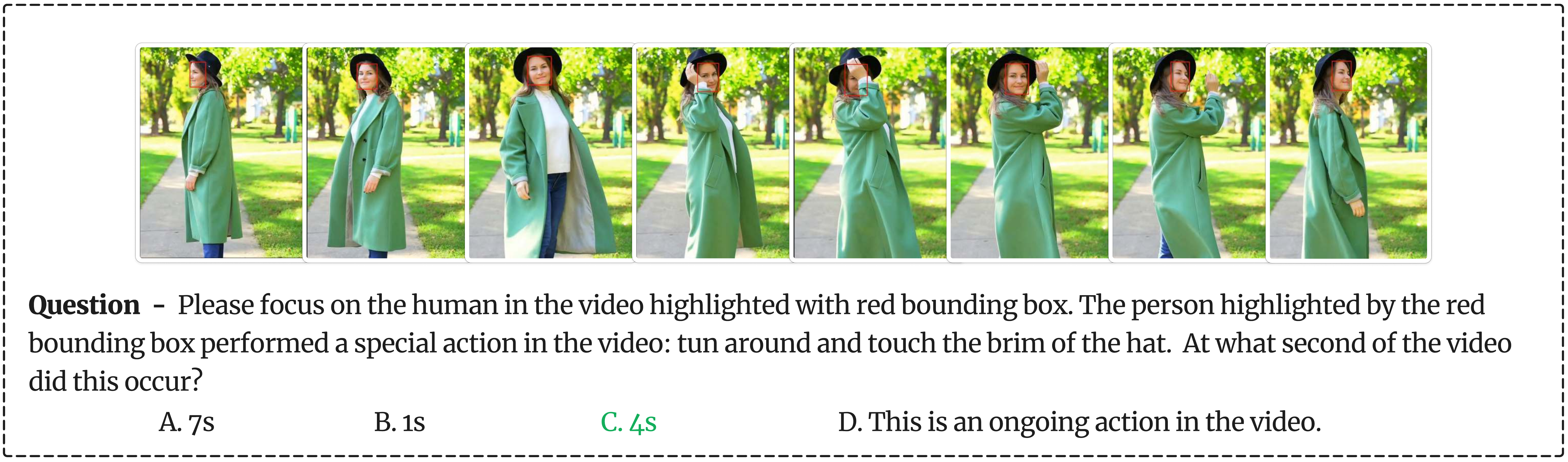}
    \vspace{-0.5em}
    \caption{Example of Time of Specific Action task.}
    \label{fig:Example_of_Time_of_Specific_Action}
    \vspace{-1.0em}
\end{figure}

\noindent \textbf{Time of Specific Action} focuses on determining the time when a specific behavior occurs, requiring the model to accurately pinpoint the time of a particular action in the video. An example is shown in Figure \ref{fig:Example_of_Time_of_Specific_Action}.

\subsection{Cross-Modal Speech-Visual Alignment}
\begin{figure}[h!]
    \centering
    \includegraphics[width=1\columnwidth]{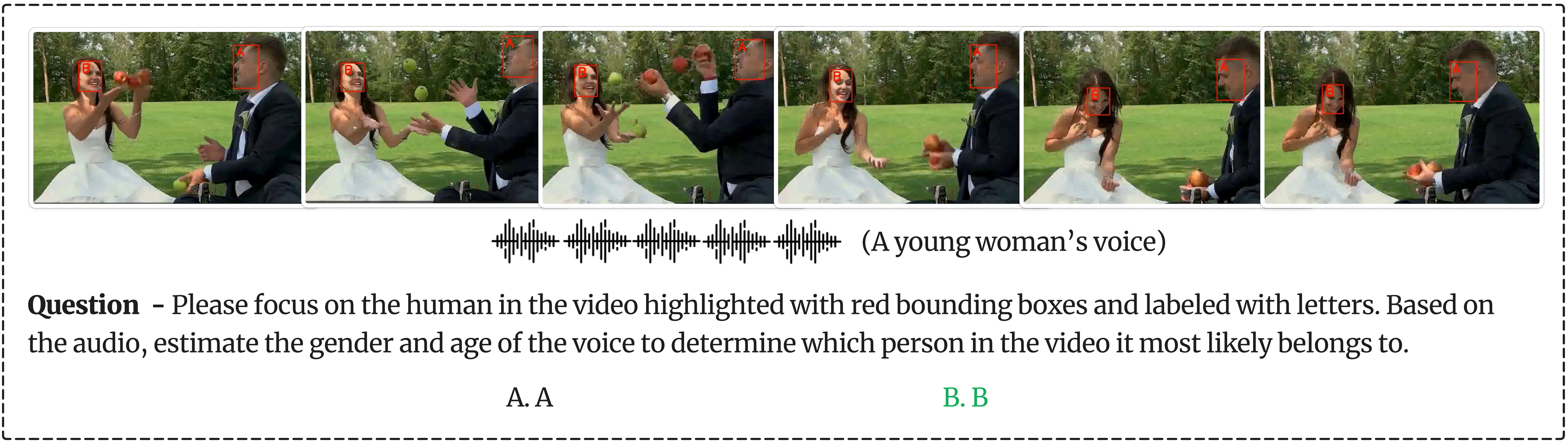}
    \vspace{-0.5em}
    \caption{Example of Audio-Visual Speaker Matching task.}
    \label{fig:Example_of_Audio_Visual_Speaker_Matching}
    \vspace{-1.0em}
\end{figure}

\noindent involves analyzing audio cues in multi-person videos to identify the individual whose appearance matches the voice. This task evaluates whether the model can recognize the voice gender and age and compare them with the appearance of the person in the video. An example is shown in Figure \ref{fig:Example_of_Audio_Visual_Speaker_Matching}.\\

\begin{figure}[h!]
    \centering
    \includegraphics[width=1\columnwidth]{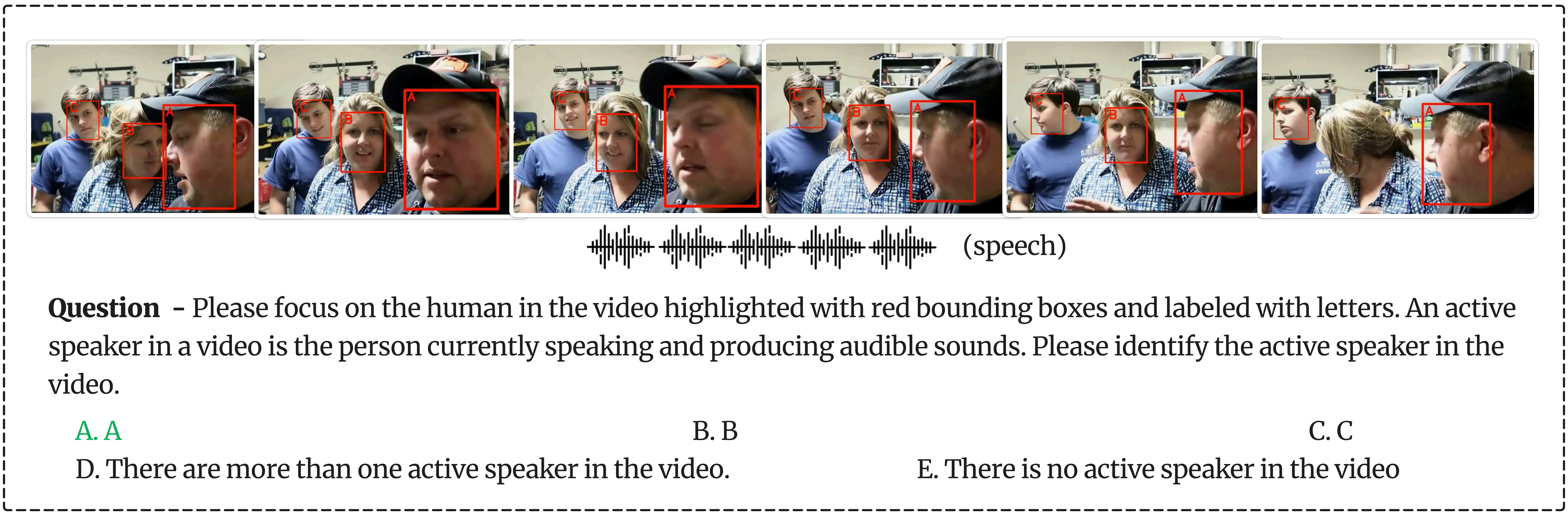}
    \vspace{-0.5em}
    \caption{Example of Active Speaker Detection task.}
    \label{fig:Example_of_Active_Speaker_Detection}
    \vspace{-1.0em}
\end{figure}

\noindent \textbf{Active Speaker Detection} asks the model to identify the active speaker in the video, requiring the model to accurately identify who is speaking by combining audio cues with the characters' lip movements. An example is shown in Figure \ref{fig:Example_of_Active_Speaker_Detection}.\\

\begin{figure}[h!]
    \centering
    \includegraphics[width=1\columnwidth]{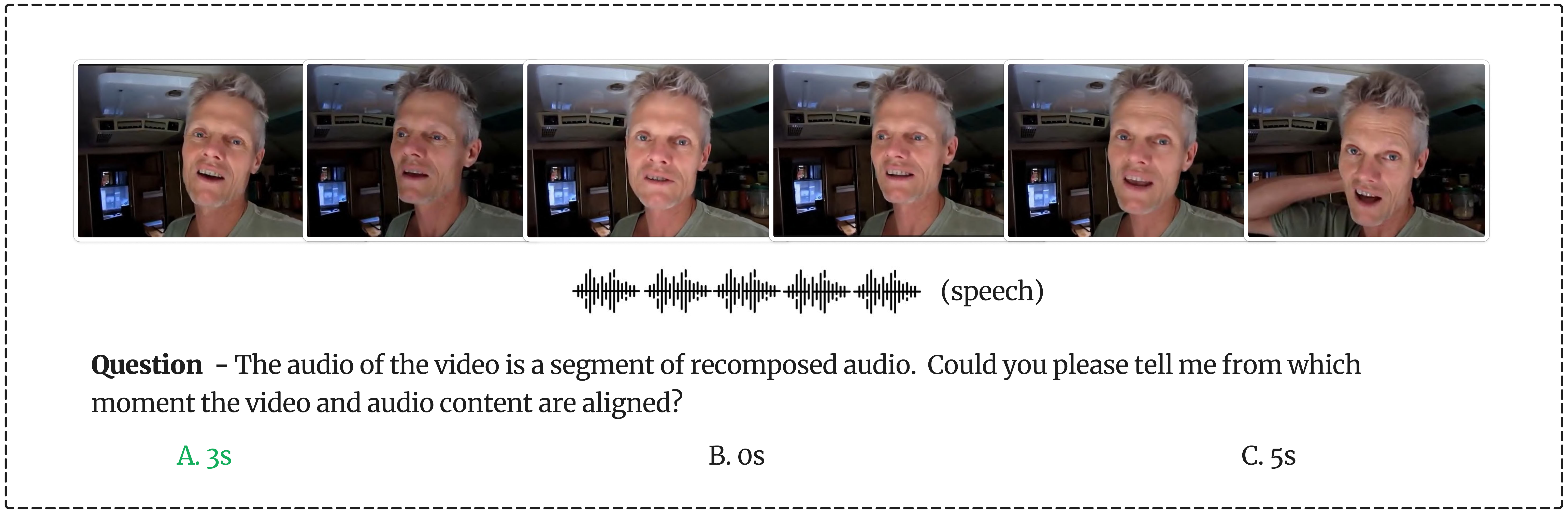}
    \vspace{-1.0em}
    \caption{Example of Audio-Visual Alignment Detection task.}
    \label{fig:Example_of_Audio-Visual_Alighment_Detection}
\end{figure}

\noindent \textbf{Audio-Visual Alignment Detection} requires detecting when the audio and video are synchronized, evaluating the model's ability to synchronize audio and visual content, particularly through analyzing the speaker's lip movements and voice. An example is shown in Figure \ref{fig:Example_of_Audio-Visual_Alighment_Detection}.\\

\begin{figure}[h!]
    \centering
    \includegraphics[width=1\columnwidth]{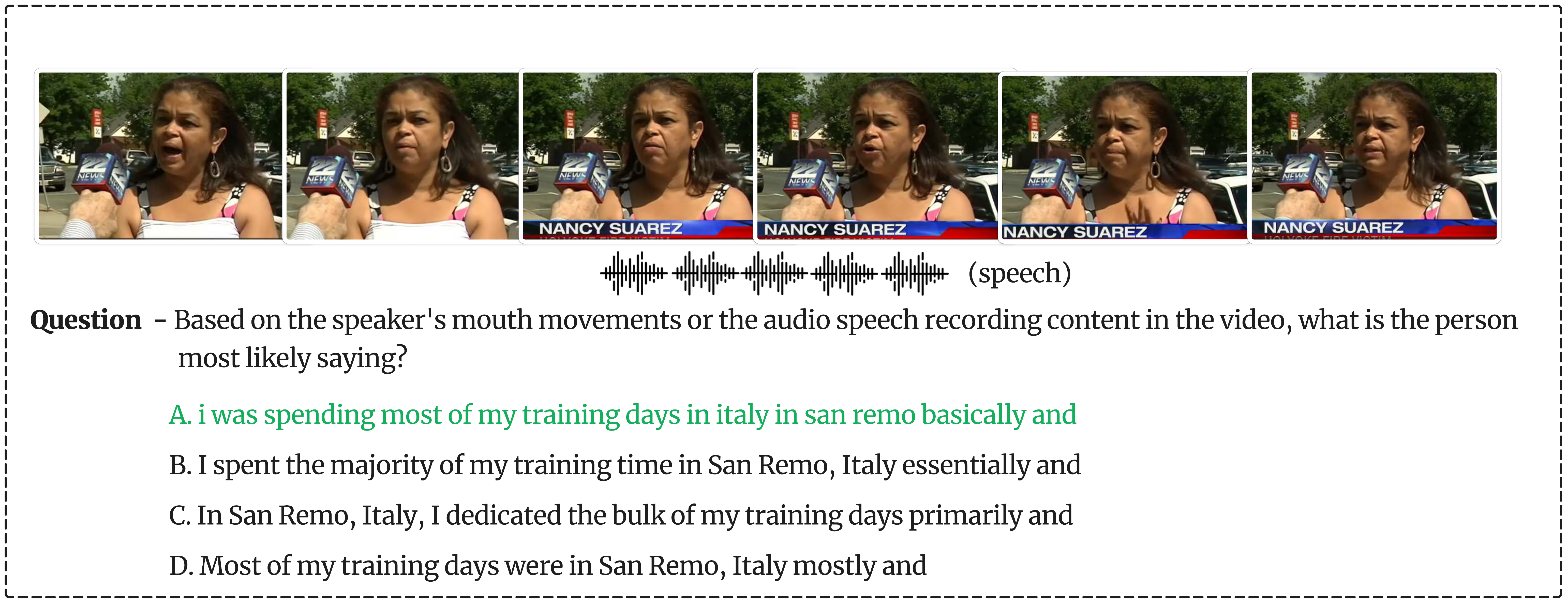}
    \vspace{-0.5em}
    \caption{Example of Speech Content Matching task.}
    \label{fig:Example_of_Speech_Content_Matching}
    \vspace{-1.0em}
\end{figure}

\noindent \textbf{Speech Content Matching} requires matching the speech content of the video with text, validating the model's ability to transcribe speech or translate lip movements into text. Figure \ref{fig:Example_of_Speech_Content_Matching} shows an example.

\section{Annotations Details and Examples in \textit{Human-Centric Annotation Pipeline}}
\label{append:annotation_pipeline}
For the in-the-wild videos collected from Pexels, we first apply splitting and filtering operations. Specifically, we begin by utilizing the {\small\texttt{video\_resolution\_filter}}, {\small\texttt{video\_aesthetics\_filter}}, and {\small\texttt{video\_nsfw\_filter}} operators to select videos that meet the following criteria: a resolution of at least 1280 in width and 480 in height, acceptable aesthetics, and appropriate content. Next, the {\small\texttt{video\_split\_by\_scene\_mapper}} is used to split the videos into scenes. The resulting clips are then filtered using the {\small\texttt{video\_duration\_filter}} to exclude clips shorter than 1 second and the {\small\texttt{video\_motion\_score\_filter}} to remove static videos. These steps utilize existing operators in Data-Juicer \citep{dj2}, with parameters set to their default values except for the {\small\texttt{video\_motion\_score\_filter}}, where the minimum motion score is set to 1.2. After completing these foundational steps, we apply the {\small\texttt{video\_face\_ratio\_filter}} with a threshold of 0.65 to retain videos containing people. These videos are then processed using a series of mappers to generate fine-grained, multi-modal, human-related annotations.

\begin{figure*}[t!]
\begin{center}
\centerline{\includegraphics[width=0.9\linewidth]{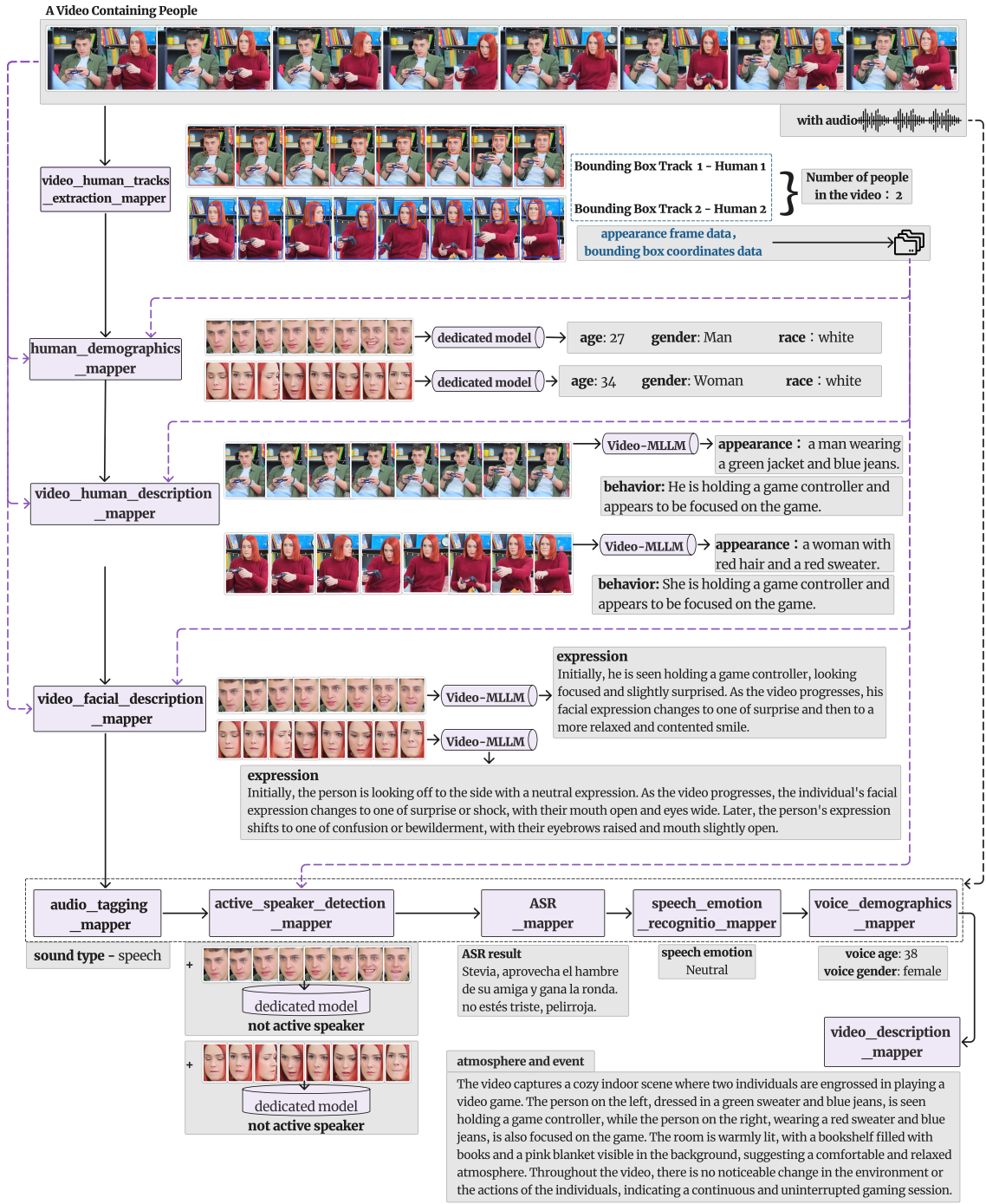}}
\caption{An example of using Human-Centric Annotation Pipeline for annotation.}
\label{fig:annotation_example}
\end{center}
\end{figure*}

We use a video example to demonstrate the annotation process and results, as shown in Figure \ref{fig:annotation_example}. Below, we detail the models and settings used for each operator. 

For the {\small\texttt{video\_human\_tracks\_extraction\_mapper}}, we follow the approach of Light\_ASD \citep{Liao_2023_CVPR}, utilizing S3FD \citep{zhang2017s3fd} as the face detector. A face bounding box is added to a human track if its overlap rate exceeds 50\%.
After obtaining the face track, we identify the corresponding body bounding box for each face bounding box in the same frame to generate a second bounding box track for the individual, referred to as the body track. The matching criterion selects the candidate bounding box with the smallest horizontal center distance and a smaller area. This process can be expressed by the following formula:

\begin{equation}
\begin{split}
    &\text{closest\_bbox} = \mathop{\arg\min}_{\text{bbox} \in \text{candidate\_bboxes}} \\
    &\left( \left( \frac{x_1 + x_2}{2} -  \frac{\text{f\_x1} + \text{f\_x2}}{2} \right)^2 + (x_2 - x_1)(y_2 - y_1) \right),
\end{split}
\end{equation}
where f\_x1,f\_x2 are the left and right boundaries of the face bounding box, the candidate human bounding boxes are obtained using  \href{https://github.com/jahongir7174/YOLOv8-human}{YOLOv8-human}, $x_1$, $x_2$, $y_1$, $y_2$ are the boundary values of a candidate human bounding box. If no bounding box meets the criteria, the frame is skipped, and detection proceeds with the other frames. Finally, the empty elements in the body track are replaced with the average of the bounding boxes from the surrounding frames.

In the {\small\texttt{human\_demographics\_mapper}}, we use \href{https://viso.ai/computer-vision/deepface/}{DeepFace} to perform frame-level detection of facial gender, age, and race. The analysis is conducted on cropped frames obtained directly from the {\small\texttt{video\_human\_tracks\_extraction\_mapper}} results. Finally, for a given face track, the demographics features are determined by taking the mode of the frame-level gender and ethnicity detections, and the median of the age detections.

In the {\small\texttt{video\_human\_description\_mapper}}, we use the body bounding box track to crop the video, creating a reconstructed video focused on a single individual. This reconstructed video is then processed using ShareGPT4Video \citep{chen2024sharegpt4video} for appearance description and simple actions.

In the {\small\texttt{video\_facial\_description\_mapper}}, we use the face bounding box track to crop the video, creating face-focused reconstructed videos for emotion description using Qwen2.5-VL \citep{cheng2024videollama}. 

The {\small\texttt{audio\_tagging\_mapper}} is a built-in operator in Data-Juicer, which we use directly for audio type classification.

The core model for the operator {\small\texttt{active\_speaker\_detection\_mapper}} is ASD-Light \citep{Liao_2023_CVPR}. Each face track sequence is analyzed together with the corresponding audio segment for the same time period. The model outputs a score sequence of the same length as the face track's frames, where each score evaluates whether the individual is speaking in the current frame. Positive scores indicate active speaking, while negative scores indicate not. To assign a binary ``speak or not'' label to a human track, we classify an individual as an active speaker if the longest sequence of consecutive positive scores exceeds 12 frames. Notably, to reduce false positives, we cross-check the voice-based gender and age attributes with the individual's demographic features. If there is a significant mismatch, the positive label is reassigned as negative.

The automatic speech recognition model used in the {\small\texttt{ASR\_mapper}} is SenseVoice \citep{an2024funaudiollmvoiceunderstandinggeneration}, which can also be utilized in the {\small\texttt{speech\_emotion\_recognition\_mapper}}. For the {\small\texttt{voice\_demographics\_mapper}}, we use the wav2vec2 \citep{baevski2020wav2vec} model.

The results of the {\small\texttt{video\_description\_mapper}} are not directly involved in the construction of multiple-choice questions in this work. However, the environment, atmosphere, and events occurring in the video play a crucial role in understanding the actions and expressions of individuals. Therefore, we have included this mapper in the Human-Centric Annotation Pipeline. The example shown in Figure \ref{fig:annotation_example} is generated by ShareGPT4Video.

Notably, in the Human-Centric Video Annotation Pipeline, all the models we use are based on the most advanced open-source models available. As more powerful and specialized models emerge, integrating them into our pipeline can further enhance the quality of annotations. 

\section{Complete Construction Details of All Tasks}
\label{append:annotation_details_all_task}
We will first explain the details of six descriptive questions generated using the Distractor-Included QA Generation Pipeline, followed by the construction details of the remaining tasks.

\subsection{Construction Details of 6 Descriptive Human-Centric Questions}
We first present the general instruction templates in the six task generation processes.

The prompt template for the Video-MLLMs used to obtain an answer based on the question is:
\textcolor{gray}{\textit{Answer the following question based on the video. Respond in one concise sentence. \textlangle question\_text\textrangle}}.
The obtained answer may later serve as the ensemble best answer or as material for distractors.

The prompt used to rank all candidate answers is: \\ \textcolor{gray}{\textit{Task: Given the following question and its four answer options, rank the options in order of accuracy from highest to lowest. Question: \textlangle question\textrangle, Options: answer1: \textlangle answers[0]\textrangle, answer2: \textlangle answers[1]\textrangle, answer3: \textlangle answers[2]\textrangle, answer4: \textlangle answers[3]\textrangle. Please return the ranking in the following format: [(``answer1'', ``Answer content'', Accuracy score), (``answer2'', ``Answer content'', Accuracy score), (``answer3'', ``Answer content'', Accuracy score), (``answer4'', ``Answer content'', Accuracy score)].}} We then aggregate the accuracy scores of each answer across all MLLMs. Accounting for model capability differences, we assign different weights; in our setup, Gemini's rankings receive twice the weight of the others. The answer with the highest total score is treated as the temporary correct option, while the remaining answers serve as material for constructing distractors.

The prompt template used in the ``LLM for Generating Distractors'' in Figure \ref{fig:MCQ Generation Pipeline} is:\\
\textcolor{gray}{\textit {Below is a ready-made question and its multiple-choice options: \textlangle Question\textrangle, Proper Answer: \textlangle Answer\textrangle, Distractors1: \textlangle 
eliminator1\textrangle, Distractors2: \textlangle 
eliminator2\textrangle, Distractors3: \textlangle 
eliminator3\textrangle. This question-option set may have the following issue: The current distractors have no errors; they simply represent alternative answers to the question. This makes the correct answer less distinct compared to the distractors. Therefore, I would like your help to add minor, distinct errors to each distractor so that the correct answer is clearly the only Proper Answer. Here are the minor errors type available for selection: \textlangle error type\textrangle. \\
Remember that the modified distractors must meet the following requirements: 1. Be modified from the original distractor with only slight changes. You are not allow to creat new ones from scratch. 2. Be distinctly different from the Answer, without being overly semantically similar. Minor errors can be added. 3. Differ from each other. 4. Distractors should have similar length to the correct answer. If it is too short, lengthen the description. }}

In addition to the questions and options, the differences include the \textlangle error types\textrangle. Next, we describe the construction of each task in detail.
\\ \hspace*{\fill} \\
\textbf{Emotion Recognition}: Since \textit{Label-5} is naturally a description based on face-focused cropping videos, it is directly used as the task-oriented caption. Additionally, \textit{Label-4} is included in the task-oriented caption to enhance the detail of the questions. Considering that most in-the-wild videos exhibit positive or neutral emotions, while we aim to ensure a sufficient proportion of negative emotions in the evaluation, videos for question generation are preselected at a ratio of positive:neutral:negative = 1:1:2. This selection is achieved by using an LLM to classify the emotional polarity of the descriptions. The resulting balanced category captions are used for question generation, with the following prompt:\\
\textcolor{gray}{\textit{Please generate one question and answer pair based on the person's description: \textlangle task-oriented caption\textrangle. the question should closely related to emotion recognition. Here is an question example: ``What emotions might the girl in red dress be experiencing during her practice?''}}\\
The video for the question is marked using the face bounding box from the target character's \textit{Label-1}. The type of minor errors (\textlangle error types\textrangle) introduced for building distractors is: \textcolor{gray}{\textit{Add incorrect emotional descriptors or modify the original emotional descriptors to incorrect ones.}}
\\ \hspace*{\fill} \\
\textbf{Emotion Temporal Analysis}: We first select videos longer than 7 seconds and then identify described characters with emotional changes based on \textit{Label-5} (using an LLM for binary classification). The videos of these characters are used for question generation. For this task, \textit{Label-5} is directly used as the task-oriented caption, with \textit{Label-4} added to enhance the details of the questions. The question generation prompt is:\\
\textcolor{gray}{\textit{Please generate a question-answer pair based on the following video caption. Please note that the questions must be related to the emotional temporal changes. Here are some example: 1. How does the girl in red's emotions change as the video progresses? 2. How does the girl in red's emotions change as she dances in the video?}}\\
The video for the question is marked using the face bounding box from the target character's \textit{Label-1}. The type of minor errors introduced for building distractors is: \textcolor{gray}{\textit{Add some incorrect emotions to the sequence, remove some correct emotion words, or change the original emotional descriptors to incorrect ones.}}
\\ \hspace*{\fill} \\
\textbf{Behavior Temporal Analysis}: First, videos longer than 7 seconds are selected for question generation. Then, the target character in the video is highlighted using the face bounding box track from \textit{Label-1}. Based on the marked videos, appearance cues of the target character (i.e., \textit{Label-4}) are added to help to guide the model's attention to the individual. The prompt for obtaining the task-oriented caption is designed as follows:\\
\textcolor{gray}{\textit{Please focus on the person highlighted by the red bounding box (\textlangle Human\_Appearance\textrangle) and tell me if the actions of the person changed over time and what actions does the person take in order? Respond according to the following format: \{``Action\_Change'': True or False, ``Action\_Sequence'': action sequence\}.}}\\
Based on the task-oriented captions, select characters with changes in actions for LLM question generation. The prompt for question generation is:
\textcolor{gray}{\textit{Please generate a question-answer pair based on the following human's behavior caption. The generated questions should focus on identifying the action sequence of the highlighted person. The following is a description of a human in the video. \textlangle action\_sequence\textrangle. The focused person is \textlangle appearance\textrangle. Here is a question template you can refer to: What actions and behaviors does the girl in the red dress display in the video in order? List them sequentially. Remember do not reveal the answers in your questions and the answer should be brief and just in one sentence.}}\\
The type of minor errors introduced for building distractors is: \textcolor{gray}{\textit{Add some nonexistent actions, remove some actions, or replace correct actions with incorrect ones.}}
\\ \hspace*{\fill} \\
\textbf{Emotion Intensity Compare}: First, count the number of frames corresponding to the track with the longest appearance time in each video. If there are more than three and less than seven tracks in the video that reach this number of frames, keep the video for question generation. This step mainly use the information from \textit{Label-1} . All individuals corresponding to the tracks with the most frames will be used for question generation. The question video is created by utilizing these human tracks to mark the individuals and adding letter labels. For this task, initial question-answer pairs are directly created. The question is, ``Which person in the image, \textlangle LETTERS\textrangle, has the strongest emotion? What specific emotion is it? Please respond briefly in the format \textlangle letter-emotion\textrangle.'',
in which \textlangle LETTERS\textrangle \quad refers to all the selectable individuals' letter labels. The answer is, ``The emotional intensity of the selectable characters in the image is similar, and they are all neutral.'' The subsequent three models will refine the answer.\\
The type of minor errors introduced for building distractors is: \textcolor{gray}{\textit{If the letters are the same, minor modifications to the emotions can be made to make the options different; if the letters referring to people are different, the emotions can remain unchanged.}}
\\ \hspace*{\fill} \\
\textbf{Human-to-Text}: First, select videos with 3 to 7 individuals based on \textit{Label-2}, and then choose the person who appears the most frames in the video as the target individual for question generation. Next, highlight the target individual in the video using the face bounding box track from \textit{Label-1}. Based on the marked video, appearance cues of the target individual (i.e., \textit{Label-4}) are added to help the model focus on the person. The prompt for obtaining the task-oriented caption is designed as follows:\\
\textcolor{gray}{\textit{Please accurately describe the person highlighted by a red box(\textlangle appearance\textrangle), your answer can be based on appearance, location, and actions, so that the highlighted person can be distinguished from others in the video. Please respond in only one sentence and begin with ``The person is ...''.}}\\
Based on the above description of the target individual, the question-answer pair is directly constructed. The question is fixed as: ``Please accurately describe the person highlighted by a red box based on appearance, location, and actions, so that the highlighted person can be distinguished from others in the video.'' The initial answer is the task-oriented caption.\\
The type of minor errors introduced for building distractors is:\textcolor{gray}{\textit{Based on the items, people, and position information, add small modifications to make the location information incorrect; alternatively, you can also modify the description of the person's appearance to introduce errors.}}
\\ \hspace*{\fill} \\
\textbf{Behavioral Causality Analysis}: The construction process is similar to the design process of Behavior Temporal Analysis. First, videos longer than 7 seconds are selected for question generation. Then, the target individual in the video is highlighted using the face bounding box track from \textit{Label-1}. Based on the annotated video, appearance cues of the target individual (i.e., \textit{Label-4}) are added to assist the model in identifying the person to focus on. The prompt for obtaining the task-oriented caption is designed as follows:\\
\textcolor{gray}{\textit{Please describe the causal events related to the person highlighted by the red bounding box (\textlangle appearance\textrangle) in the video: what causes this person to exhibit a certain behavior, or what actions does this person take that led to a certain event. If no causal events exist, respond without causal events. Please answer in the following format: \{``causal\_events\_exist'': True or False, ``causal\_events\_description'': description\}.}}\\
Videos and target individuals with causal relationships (i.e. ``causal\_events\_exist'' is true) are then selected for question generation. The prompt for question generation is:
\textcolor{gray}{\textit{Please generate a question-answer pair based on the following video caption. The generated questions should inquire about causal reasoning related to the character's expressions or behaviors. The following is a description of the human in the video: \textlangle causal\_events\_description\textrangle. The focused person is \textlangle appearance\textrangle . You should either follow the causal analysis question template ``Analyze why the girl in the red dress raises her hand.'' or the result derivation question template ``What does the girl in the red dress raising her hand lead to?''. Remember do not reveal the answers in your questions and the answer should be brief and just in one sentence.}}\\
The type of minor errors introduced for building distractors is: \textcolor{gray}{\textit{Explain the result using incorrect causes, misdescribe the effect of the cause-and-effect relationship, reverse the order of cause and effect, exaggerate or minimize factors.}}

\subsection{Construction Details of 10 Closed-Ended Human-Centric Questions}
\textbf{Attitude Recognition}: This task is constructed based on the first half of the Distractor-Included QA Synthesis Pipeline. The human who appears in the most frames is selected as the target for question generation. The target individual is highlighted in the video using the face bounding box track from \textit{Label-1}. Based on the annotated video, appearance cues of the target individual (i.e., \textit{Label-4}) are added to the prompt to help the model focus on the intended person. The prompt used to obtain the task-specific caption is:\\
\textcolor{gray}{\textit{Focus on the person highlighted by the red bounding box (\textlangle appearance\textrangle) and tell me: Do the highlighted people display certain attitudes toward specific objects and events? What kind of attitude is it?}}\\
The prompt used for question generation is:\\
\textcolor{gray}{\textit{Please generate a best question-answer pair based on the following video caption. The generated questions should focus on analyzing the character's attitude, which should be one of positive, negative, or neutral. The following is a description of the human in the video. \textlangle task-specific caption\textrangle The focused person is \textlangle appearance\textrangle. Here are some question templates you can refer to: 1. What is the attitude of the girl in the blue shirt towards taking the bus in the video? positive, negative, or neutral? 2. What is the woman in the beige jacket's attitude? Positive, negative, neutral? Please remember not to reveal the answers in your questions and the answer should be brief and just in one sentence. }}\\
The options consist of four choices: Positive, Negative, Neutral, and Indeterminate. The Indeterminate option is included as a supplemental choice to ensure answers optional.
\\ \hspace*{\fill} \\
\\ \hspace*{\fill} \\
\textbf{Text-to-Human}: The criteria for selecting the videos for questioning are consistent with the Emotion Intensity Compare task selection rules. Then, use the same method as Human-to-Text to obtain task-specific captions and directly use the description of the target person to complete the question template: ``Please select the person in the video that best matches the following description: \textlangle human description \textrangle''. The video corresponding to the question is marked with the face bounding boxes of all individuals using \textit{Label-1}, and each individual is distinguished by a capital letter label. The selectable options are the letter labels representing each person.
\\ \hspace*{\fill} \\
\textbf{Human Counting}: For an annotated video, the approximate number of people in the video can be estimated directly using \textit{Label-2}. However, due to issues such as blurred crowd background, overlapping between people and objects and other factors, this estimate is often imprecise, especially in crowded scenes. Therefore, \textit{Label-2} is only used to adjust the question distribution (3–5 people: 60, 6–8 people: 60, 9+: 54). The ground truth number is manually annotated, and distractors are constructed based on this value. The  distractors construction rule is to randomly select three different numbers within a range of up to 4 from the ground truth number, excluding the ground truth itself. 
\\ \hspace*{\fill} \\
\textbf{Appearance Time Detection}: First, select videos based on the following criteria: the video duration must exceed 7 seconds, and the target individual's presence should account for between one-third and two-thirds of the total video length (calculated as the ratio of the human track frames to the total frames), primarily using \textit{Label-1}. Then the frame range from \textit{Label-1} is used to determine the target individual's appearance time range (format both ends as integers), which serves as the ground truth for generating questions about this person.

To obtain a detailed and accurate description of the individual, the same method as in the Human-to-Text task is used to generate the task-specific caption for the target. Using the description, questions are constructed in a template-based manner, as shown in Figure \ref{fig:Example_of_Appearance_Time_Detection}. 

For distractor construction, three random time intervals are generated near the ground truth time interval, ensuring that their overlap with the ground truth interval does not exceed 4 seconds. This ensures the distractors do not cause confusion when selecting the correct answer.

Note that in this task, videos with bounding boxes are only used during the automatic description generation by Video-MLLMs and for manual verification. In the final version of the questions, the videos do not include bounding boxes.
\\ \hspace*{\fill} \\
\textbf{Action at Specified Time} and \textbf{Time of Specific Action} tasks rely on manual annotation, as attempts with various open-source models revealed their inability to accurately identify the timing of specified actions. For manual annotation, annotators are required to watch the videos and observe whether the highlighted individual performs any distinct short-term actions (quickly completed actions or  ``the start of an action'', but not continuous states). They should record the action and its starting time. 
The videos and target individuals are consistent with those in Behavior Temporal Analysis task. Based on the specific action–time pairs provided by the annotators, two types of action-time-related questions are constructed.

\begin{table*}[h]
\belowrulesep=0pt
\aboverulesep=0pt
\centering
\renewcommand{\arraystretch}{1.0} 
\scriptsize 
\setlength{\tabcolsep}{3pt} 
\begin{tabular}{l|llll|llll|llll|llll}
\toprule
 & \multicolumn{4}{c|}{\textbf{Human Emotion Perception}} & \multicolumn{4}{c|}{\textbf{Person Recognition}} & \multicolumn{4}{c|}{\textbf{Human Behavior Analysis}} & \multicolumn{4}{c}{\textbf{Speech-Visual Alignment}} \\

 \textbf{Labels} & \textbf{ER} & \textbf{ETA} & \textbf{AR} & \textbf{EIC}& \textbf{T2H} & \textbf{H2T} & \textbf{HC} & \textbf{ATD} & \textbf{BTA} & \textbf{BCA} & \textbf{AST} & \textbf{TSA}& \textbf{AVSM} & \textbf{ASD} & \textbf{AVAD} & \textbf{SCM}\\
\toprule
\textbf{\textit{Label-1}} & \ding{51}&\ding{51}&\ding{51}&\ding{51}&\ding{51}&\ding{51}& &\ding{51}&\ding{51}&\ding{51}&\ding{51}&\ding{51}&\ding{51}&\ding{51}& & \\
\hline
\textbf{\textit{Label-2}} & & & & &\ding{51}&\ding{51}&\ding{51}& & & & & &\ding{51}&\ding{51}&\ding{51}&\ding{51}\\
\hline
\textbf{\textit{Label-3}} & & & & & & & & & & & & &\ding{51}& & & \\
\hline
\textbf{\textit{Label-4}} & \ding{51}&\ding{51}&\ding{51}& &\ding{51}&\ding{51}& &\ding{51}&\ding{51}&\ding{51}&\ding{51}& & & & \\
\hline
\textbf{\textit{Label-5}} & \ding{51}&\ding{51}& & & & & & & & & & & & & & \\
\hline
\textbf{\textit{Label-6}} & & & & & & & & & & & & &\ding{51}&\ding{51}&\ding{51}&\ding{51}\\
\hline
\textbf{\textit{Label-7}} & & & & & & & & & & & & & &\ding{51}&\ding{51}&\ding{51}\\
\hline
\textbf{\textit{Label-8}} & & & & & & & & & & & & & & & &\ding{51}\\
\hline
\textbf{\textit{Label-10}} & & & & & & & & & & & & &\ding{51}& & &\\

\bottomrule
\end{tabular}
\caption{Annotation labels used in the construction process of 16 tasks} 
\label{table:label-task relation}
\end{table*}

For the Action at Specified Time task, the question video consists of the highlighted target individuals with red bounding boxes. Only video samples where short-term actions are present are selected for question generation. The question template is shown in Figure \ref{fig:Example_of_Action_at_Specific_Time}. The ground truth is the specific action annotated for the individual. Distractors are generated by LLM from the task-specific captions in Behavior Temporal Analysis, with the following prompt:\\
\textcolor{gray}{\textit{Please select and modify 3 actions from the list below to ensure that each action is significantly different from the target action \textlangle ground truth action\textrangle. Here is the original action list: \textlangle action\_list\textrangle. Begin with ``begin to ..'' for an action. If the number of actions is less than 3, generate one.}},\\
where the \textlangle action\_list\textrangle is the sequence of actions generated by using LLM to summarize the task-oriented caption.

For the Time of Specific Action task, the question video is marked with target individual's bounding boxes. The question template is shown in Figure \ref{fig:Example_of_Time_of_Specific_Action}. For question samples with short-term actions, distractors are generated by selecting three numbers that are at least 3 seconds apart from the ground truth action time. For videos where a continuous action state is maintained throughout, the ground truth is set to ``This is an ongoing action in the video.'' The distractors for such cases are fixed at 1s, 4s, 7s, and 10s. Note that to keep the options consistent, each question includes the option ``This is an ongoing action in the video.''
\\ \hspace*{\fill} \\
\textbf{Audio-Visual Speaker Matching}: First, select the appropriate question videos. The constraints mainly include video conditions and character conditions. The video conditions include: the audio label being ``Speech'', the number of people in the video is between 2 and 4, and the video duration is not less than 4 seconds; the character condition is: the frame coverage of the character must reach more than 67\% of the video frame number. Further, the target person is selected as the ground truth according to the correlation between the age and gender attributes of the audio and the appearance of the person. Specifically, if the audio age belongs to ``child'', the only child is selected from the video as the target person, and the other characters are interference characters; If the audio age is ``adult'', the only adult with the same gender as the audio is selected from the video as the target person, and the other characters are interference characters. The age information is binary here because the audio age attribute is relatively vague. In addition, the gender characteristics of children's voices are not always distinguishable. Therefore, in order to enhance the optionality of the answer, the character types are divided into only three categories: male, female, and child. The question video uses \textit{Label-1} to mark each optional person and capital letters as the option.
\\ \hspace*{\fill} \\
\textbf{Active Speaker Detection}: Suitable question videos are selected based on the criteria of having an audio label of ``Speech'', 2 to 4 people, and a duration of at least 4 seconds. The video must contain a single active speaker. \textit{Label-1} is used to label all individuals, with the active speaker’s label as the ground truth and others as distractors. Since the automated active speaker labels may not always be reliable, two additional options are included for each question to facilitate manual correction later: ``There are more than one active speaker in the video.'' and ``There is no active speaker in the video.''
\\ \hspace*{\fill} \\
\textbf{Audio-Visual Alignment Detection}: Suitable question videos are selected based on the criteria of having fewer than 3 people, a duration of over 8 seconds, an audio type of ``speech'', and at least one active speaker. The video is then divided into three equal segments, with the left endpoint of each segment (rounded to an integer) used as potential options. One of these options is randomly chosen as the ground truth. The video is then modified by reversing the audio before the selected timestamp to create a ``misaligned audio-visual'' video.
\\ \hspace*{\fill} \\
\textbf{Speech Content Matching}: Videos are selected for question creation based on the following criteria: single-person scenes, duration greater than 5s, audio type ``speech'', and the person is an active speaker, the speech content being English with its sentence length greater than 35 characters. The ground truth is the automatic speech recognition result corresponding to \textit{Label-8}. Distractors are generated using an LLM, which creates three different sentences with similar meaning and length to the ground truth as distractors.
\\ \hspace*{\fill} \\
In Table \ref{table:label-task relation}, we illustrate which labels from the Human-Centric Video Annotation Pipeline are used to construct each task.

\subsection{Details of Human Efforts for \ours}
We employed two professional full-time AI data annotators and invited two graduated-level volunteers to participate in the construction and evaluation of \ours. They collaborate to complete a series of tasks, with two main annotators participating in the full data of each task and two volunteers participating in the sampled data. Inconsistencies will be identified and resolved (for example, through discussion or majority voting to reach a consensus) to ensure high quality. The average annotation time per question is 3 minutes, totaling 10 workdays.
Specifically, their tasks include the following four parts.
\\ \hspace*{\fill} \\
\textbf{Human Annotation on Generated QAs}: Tasks requiring human annotations to generate questions and options include Human Counting, Action at Specified Time, and Time of Specific Action. The latter two tasks can streamline annotation by annotating a single dataset containing ``special short-term action \& moment of occurrence''. Therefore, in this step, each annotator was assigned to one annotation set. All other tasks were generated automatically, reducing the cost of human annotation.
\\ \hspace*{\fill} \\
\textbf{Manual Verification and Correction}: Except for the tasks that are already reliable enough, which include three tasks derived from the aforementioned human annotations, the Audio-Visual Alignment Detection and Speech Content Matching tasks, all other tasks require manual verification to ensure quality, following the process described in Section \ref{main:Manual Verification}. During correction, low-quality samples (e.g., person transitions in human tracking, video freezing midway) are required to be flagged for removal. 
\\ \hspace*{\fill} \\
\textbf{Cross-Verification}: After completing the above steps, we obtained 16 usable tasks for evaluation. To further ensure the high quality of the questions, we conducted cross-verification on 16 tasks except Emotion Recognition in Conversaton task  to reduce the impact of personal biases and errors on the benchmark. 
Specifically, the tasks were cross-assigned to the two major annotators. For each task, the annotator in this step was ensured to be different from those responsible for the Manual Verification and Correction or Human Annotation. Annotators were first required to answer the multiple-choice questions. For disputed questions where answers were marked ``incorrect'', the new correct answer or option will be updated for this question, and these disputed questions were reassigned to another annotator for a second review. If errors persisted, both annotators discussed and agreed on a unified answer to serve as the final ground truth.

\end{document}